\documentclass[letterpaper]{article} %
\usepackage[submission]{aaai24}  %
\usepackage{times}  %
\usepackage{helvet}  %
\usepackage{courier}  %
\usepackage[hyphens]{url}  %
\usepackage{graphicx} %
\usepackage{natbib}  %
\usepackage{caption} %
\usepackage{algorithm}

\usepackage{newfloat}
\usepackage{listings}
\DeclareCaptionStyle{ruled}{labelfont=normalfont,labelsep=colon,strut=off} %
\floatstyle{ruled}
\newfloat{listing}{tb}{lst}{}
\floatname{listing}{Listing}
\usepackage[utf8]{inputenc} %
\usepackage[T1]{fontenc}    %
\usepackage{url}            %
\usepackage{booktabs}       %
\usepackage{amsfonts}       %
\usepackage{nicefrac}       %
\usepackage{microtype}      %

\title{Variational Sampling of Temporal Trajectories}

\author{
Jurijs Nazarovs\textsuperscript{\rm 1},
Zhichun Huang\textsuperscript{\rm 2},
Xingjian Zhen\textsuperscript{\rm 1},
Sourav Pal\textsuperscript{\rm 1},
Rudrasis Chakraborty\textsuperscript{\rm 3},
Vikas Singh\textsuperscript{\rm 1} \\
\textsuperscript{\rm 1}University of Wisconsin-Madison \\
\textsuperscript{\rm 2}Carnegie Mellon University \\
\textsuperscript{\rm 3}Butlr \\
{\tt\small nazarovs@wisc.edu, zhichunh@alumni.cmu.edu, zhenxingjian1995@gmail.com, spal9@wisc.edu, rudrasischa@gmail.com, vsingh@biostat.wisc.edu} \\
\url{https://github.com/vsingh-group/functional_node}
}

\usepackage[utf8]{inputenc} %
\usepackage[T1]{fontenc}    %
\usepackage{url}            %
\usepackage{booktabs}       %
\usepackage{amsfonts}       %
\usepackage{nicefrac}       %
\usepackage{microtype}      %

\usepackage[table]{xcolor}

\usepackage[american]{babel}

\usepackage{natbib} %
    
\usepackage{mathtools} %

\usepackage{tikz} %

\usepackage{url}
\usepackage{graphicx}
\usepackage{dsfont}
\usepackage{amsmath}
\usepackage{amssymb}

\usepackage{empheq}

\usepackage{graphicx}
\usepackage{subcaption}
\usepackage{listings}

\usepackage{multirow}
\usepackage{rotating}
\usepackage{multicol}
\usepackage{mathrsfs}

\usepackage{amsthm,comment}

\theoremstyle{remark}

\newcommand{\z}{\ensuremath{\boldsymbol{z}}}

\newcommand{\rulesep}{\unskip\ \vrule\ }

\usepackage{booktabs,paralist,hhline,colortbl} 

\usepackage{algorithm}
\usepackage{algorithmicx}
\usepackage{algpseudocode}

\algnewcommand\algorithmicinput{\textbf{Input:}}
\algnewcommand\algorithmicoutput{\textbf{Output:}}
\algnewcommand\Input{\item[\algorithmicinput]}%
\algnewcommand\Output{\item[\algorithmicoutput]}%

\definecolor{Maroon}{RGB}{204, 102, 0}

\definecolor{rulecolor}{rgb}{0.0, 0.06, 0.54}

\definecolor{tableheadcolor}{rgb}{0.88, 0.94, 0.87}
\definecolor{orange}{rgb}{1, 0.49, 0}
\definecolor{bluecolor}{rgb}{0.74, 0.83, 0.9}

\usepackage{tcolorbox}
\newtcolorbox{mybox}[3][]
{
  colframe = #2!15,
  colback  = #2!10,
  coltitle = #2!10!black,  
  title    = {#3},
  boxsep   = 0.25pt,
  left     = 0.5pt,
  right    = 0.5pt,
  top      = 0pt,
  bottom   = 0pt,
  width=\linewidth,
  #1,
}

\usepackage[capitalize]{cleveref}
\crefname{section}{Sec.}{Secs.}
\Crefname{section}{Section}{Sections}
\Crefname{table}{Table}{Tables}
\crefname{table}{Tab.}{Tabs.}

\begin{document}

\maketitle

\begin{abstract}

A deterministic temporal process can be determined by its trajectory, an element in the product space of (a) initial condition $\z_0 \in \mathcal{Z}$  and (b)  transition function $f: (\mathcal{Z}, \mathcal{T}) \to \mathcal{Z}$ often influenced by the control of the underlying dynamical system. Existing methods often model the transition function as a differential equation or as a recurrent neural network. Despite their effectiveness in predicting future measurements, few results have successfully established a method for sampling and statistical inference of trajectories using neural networks, partially due to constraints in the parameterization.
In this work, we introduce a mechanism to learn the distribution of trajectories by parameterizing the transition function $f$ explicitly as an element in a function space. Our framework allows efficient synthesis of novel trajectories, while also directly providing a convenient tool for inference, i.e., uncertainty estimation, likelihood evaluations and out of distribution detection for abnormal trajectories. These capabilities can have  implications for various downstream tasks, e.g., simulation and evaluation for reinforcement learning.

\end{abstract}

\section{Introduction}

Learning to predict and infer from temporal/sequential data is an important topic in machine learning, with
numerous application in many settings, including semantic video processing \citep{nilsson2018semantic} in vision and text generation/completion \citep{lewis2019bart, radford2018improving} in natural language processing. The literature has continued to evolve, from handling evenly-sampled discrete observations (e.g., LSTM \citep{hochreiter1997long}, GRU \citep{cho2014learning}) to irregularly-sampled or even continuous observations (e.g., Neural ODE \citep{chen2018neural} / SDE \citep{rubanova2019latent, tzen2019neural}). These capabilities
have been extensively used, 
from learning complex non-linear tasks such as language modeling \citep{devlin2018bert} to forecasting of physical dynamics \citep{lusch2018deep, brunton2020machine, rudy2017data}. 

Despite the mature state of various methods currently 
available for prediction tasks involving temporal sequences, 
fewer options exist that allow sampling from distribution of trajectories, similar to VAE for standard images, and can provide the confidence of their predictions. 
Existing methods, such as Neural ODEs, do facilitate the ability to sample trajectories by varying the initial conditions of the underlying differential equation. However, these methods are limited in their ability to model real-world systems where the {\em same} initial conditions can result in vastly different outcomes, (e.g. common in reinforcement learning). Also, brain development of identical twins can differ despite starting from exactly the same state  \cite{strike2022queensland}. 
To address this limitation, Bayesian versions of Neural ODEs, similar to stochastic differential equations, can account for uncertainty in trajectories. However, these methods are not designed to handle a wide variety of different trajectories with the same initial point and primarily not used as methods to sample dynamical processes.
How can we address the problem of sampling a wide variety of different trajectories with the same initial condition? In other words, given an NODE setup in \eqref{eq:node_orig}, where $f$ is NN,  can we learn a mechanism, which allows us to sample $f$ as an element from a functional space of NN?

At a high level, one inspiration of our work is a recent result 
dealing with explicit embeddings of neural network functions \citep{dupont2022data}. 
We can treat the transition function 
$f$ of the trajectory as an entity on its own 
and embed it explicitly in a Euclidean space. There, we can 
conveniently learn the distribution of embeddings which in-turn can help us learn the distribution of the trajectories. Then, we can 
directly estimate the likelihoods and even sample new trajectories, 
with little to no overhead. 
Ideally, the embedding space will also have far fewer dimensions compared to the dimension of the data, which can make the distribution easier to learn. While \citet{dupont2022data} have explored learning the distribution of data-generating neural network functions through explicit embeddings termed \textit{functa}s, our work extends such an idea to infer temporal trajectories by jointly constructing an appropriate functional embedding space for trajectories and learning a variational statistical model on top.

{\bf Contributions:} We study a simple mechanism to embed the trajectory of temporal processes and show that one can directly perform variational inference of trajectories in this embedding space. The end result is a probabilistic model that allows efficient sampling and likelihood estimation of novel trajectories, which is immediately useful for simulation and for applications that requires statistical inference, such as outlier detection. We show via experiments that the proposed framework can achieve competitive performance in prediction tasks compared with conventional temporal models, while benefiting from the 
additional capabilities described above, already baked into the formulation.

\section{Background/notation}\label{sec:background}

{\bf Notation:} 
We denote a time-varying vector with $T$ time steps as $\z=(\z_0,\ldots, \z_{T-1})$, where each time step $\z_t$ is a vector valued sample, specifically $\z_t \in \mathbf{R}^p$, where $p$ is the number of covariates/channels.

{\bf (Neural) Ordinary Differential Equations:}
We consider the ODE as $\dot{\z}_t=f(\z_t, t)$ to model transitions, where $f(\z_t, t)$ represents a change in state $\z_t$ at time step $t$, and depends on a value of a state of the process at time $t$. 
Given the initial time $t_{0}$ and target time $t$, ODEs compute the corresponding state $\z_t$ by performing the following operations:
\begin{equation}\label{eq:node_orig}
\boldsymbol{z}\left(t_{0}\right)=\z_0, \quad \boldsymbol{z}\left(t\right)=\boldsymbol{z}\left(t_{0}\right)+\int_{t_{0}}^{t} f(\z_t, t) d t
\end{equation}
Parameterized neural networks have been shown to model $f(\z_t, t)$ for complex nonlinear dynamics as in \citep{chen2018neural, kidger2020neural}. 

{\bf Learning latent representations with a VAE:}
One of our goals is to learn a latent representation of a temporal trajectory, and so we focus on variational approximation techniques. Recall that  
Variational auto-encoders (VAE) \citep{kingma2013auto} learn a probability distribution on a latent space. We can draw samples in the latent space and the decoder can generate samples in the space of observations. In practice, the parameters of the latent distribution are learned by maximizing the \textit{evidence lower bound} (ELBO) of the intractable likelihood:
\begin{equation}\label{eq:elbo}
    \log p\left(\mathbf{x}\right) \geq 
    -KL\left(q(\mathbf{z}) \| p(\mathbf{z})\right)+\mathbb{E}_{q(\mathbf{z})}\left[\log p\left(\mathbf{x}\mid \mathbf{z}\right)\right]
\end{equation}
where $\mathbf{z}$ is a sample in the latent space from the approximate posterior distribution $q(\mathbf{z})$, with a prior $p\left(\mathbf{z}\right)$, and $\mathbf{x}$ is a reconstruction of a sample (e.g., an image or temporal sample) with the likelihood $p\left(\mathbf{x} \mid \mathbf{z}\right)$. One choice for $q$ is $\mathcal{N}(\boldsymbol{\mu}, \Sigma)$, where $\boldsymbol{\mu}$ and $\Sigma$ are trainable parameters \citep{kingma2013auto}. Note, $\mathbf{z} \in \mathcal{Z}$ denotes a latent space sample and $\mathbf{x} \in \mathcal{X}$ denote an observed space sample.

\begin{figure}[bt]
\centering
\includegraphics[width=\columnwidth]{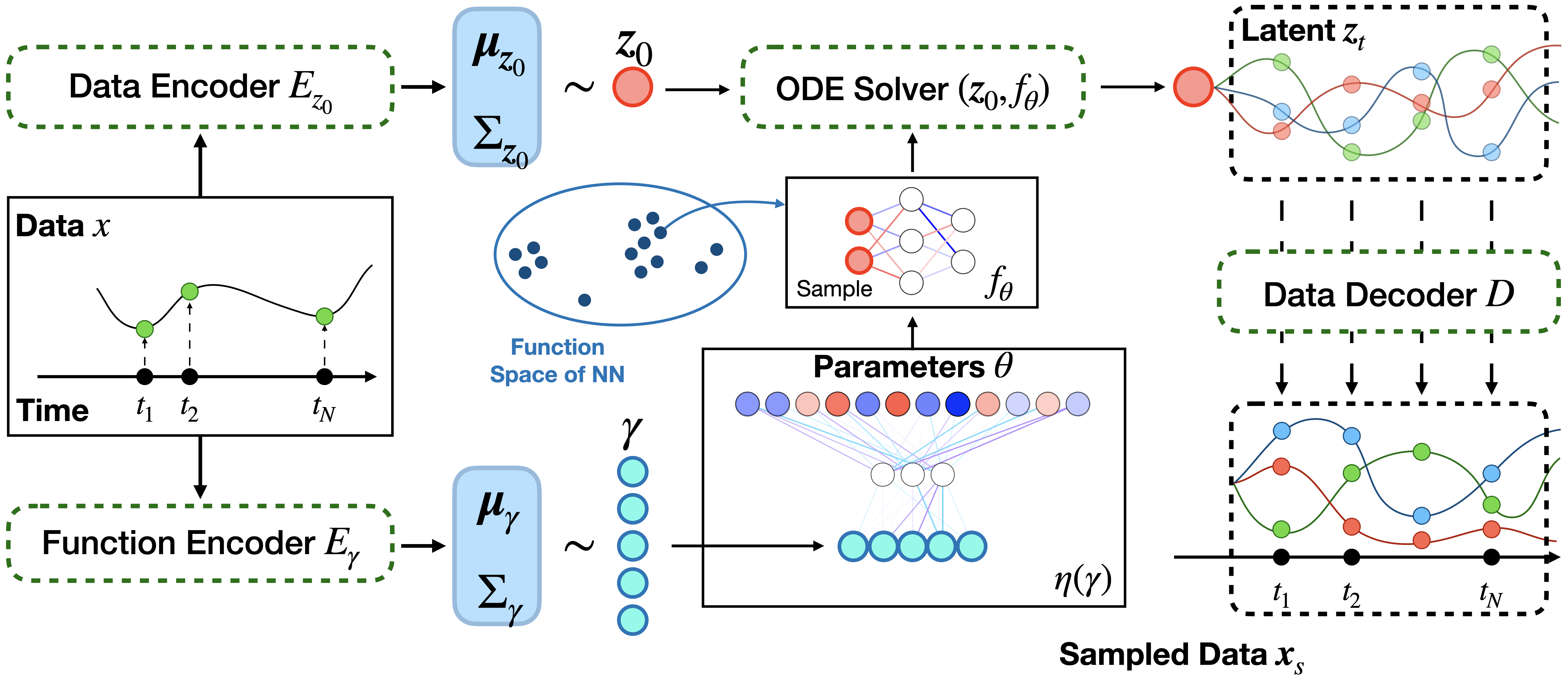} 
\caption{\footnotesize Structure of the model. First, Data Encoder is applied to temporal data sample to generate the initial point of the trajectory in latent space $\z_0$;
At the same time, the Function Encoder (embedding module) is applied to generate parameterization $\theta$ of the sample from function space, $f_\theta$. The sample from function or embedding space, $f_\theta$, is used to describe  together with a differential equation solver, to solve function DE and generate $\z_t$.
Last, decoder is used to map latent space ODE stages into observed values. 
\label{fig:model}}
\vspace{-10pt}
\end{figure}

\section{Variational Sampling of Temporal Trajectories}

 We start with deterministic temporal processes which can be defined by the corresponding trajectory on the underlying space $\mathcal{Z}$. 
A trajectory can be identified by an element in the product space of (a) initial condition $\z_0 \in \mathcal{Z}$ (often given) and (b)  transition function $f: \mathcal{Z}\times [0, T] \to \mathcal{Z}$ often influenced by the control of the underlying dynamical system.
Thus, to learn the distribution of the temporal process, it is necessary to learn distributions over both the initial condition $\z_0$
and the transition function $f$.

One way to model transition function between states $\z_t$ of the temporal process is by using the ODE: $\dot{\z}_t=f(\z_t, t)$.
We know that one can parameterize $f(\z_t, t)$ by a neural network (NN) $f_\theta(\z_t, t)$ to leverage their function approximation capacity \citep{chen2018neural}.
Thus, we can (possibly) use a NN to describe a trajectory. {\it However, if we want to sample from this space of trajectories using the above NN parameterization, we must learn a latent representation of a NN, analogous to INR \cite{dupont2022data}.} 
\vspace{-5pt}
\subsection{Functional Neural ODE}

\begin{figure}
\centering
\vspace{0pt}
\includegraphics[angle=0, width=0.15\textwidth]{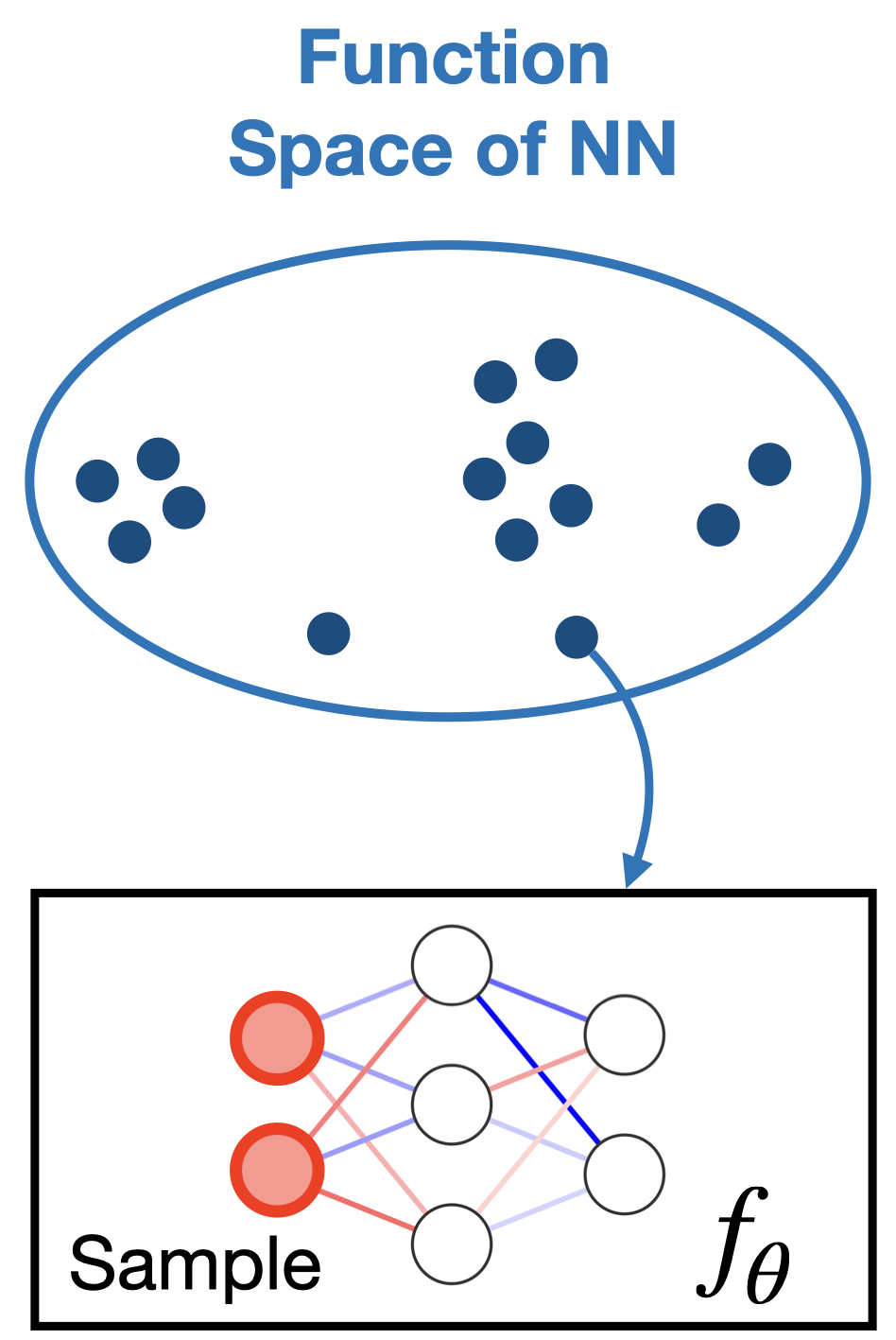} 
\caption{\footnotesize   Function/embedding space, where each element is a DNN.
\vspace{-20pt}
\label{fig:sample_net}}
\end{figure}

{\bf Goal:} Given a dataset of $N$ temporal observations with length $T$, $\{\boldsymbol{x}^{(i)}\}_{i \in [N]} = \{\boldsymbol{x}^{(i)}_0, \dots, \boldsymbol{x}^{(i)}_{T - 1}\}_{i \in [N]}$, we seek to learn a set of parameters $(\phi_{E}, \phi_{D}, \phi_{\gamma})$ corresponding to data encoder/decoder $(E_{\z_0}, D)$ and function encoder $E_{\gamma}$ (as shown in  Figure~\ref{fig:sample_net}) such that $p(\boldsymbol{x}^{(i)})$ is maximized. Formally, we can solve the following problem
\vspace{-5pt}
\begin{equation}
\max \dfrac{1}{N} \sum_{i = 1}^N p(\boldsymbol{x}^{(i)}) \equiv \max_{\phi_{D}} \dfrac{1}{N}\sum_{i = 1}^N p_{\phi_D}(\boldsymbol{x}^{(i)} | \boldsymbol{z}_0, f_\theta)p(\boldsymbol{z}_0, \theta)
\end{equation}
\vspace{-2pt}
where $\boldsymbol{z}_0$ is the initial condition, and $f_\theta$ is a NN with weights $\theta$ inferred from the data. Similar to VAE, since we do not know the true posterior distribution $p(\z_0, \theta)$, we replace it with a variational distribution $p(\boldsymbol{z}_0, \theta) \approx q_{\phi_E}(\boldsymbol{z}_0 | \boldsymbol{x}^{(i)})q_{\phi_\gamma}(\theta | \boldsymbol{x}^{(i)})$. The resultant objective is
\vspace{-5pt}
\begin{equation}
\max_{\phi_{E}, \phi{D}, \phi{\gamma}} \dfrac{1}{N}\sum_{i = 1}^N p_{\phi_D}(\boldsymbol{x}^{(i)} | \boldsymbol{z}_0, f_\theta)q_{\phi_E}(\boldsymbol{z}_0 | \boldsymbol{x}^{(i)})q_{\phi_\gamma}(\theta | \boldsymbol{x}^{(i)})
\end{equation}

\paragraph{Sampling the transition function $f_\theta$:}
In contrast to a typical Neural ODEs, we do not tie $f_\mathbf{\theta}$ to a specific parameter $\theta \in \Theta$, where $\Theta$ is the space of parameters for $f_\theta$. 
Instead, we model $\theta$ as a random variable that conditionally depends on $\boldsymbol{x}_0$. Since $\theta$ consists of all parameters in $f_\theta$ and therefore is high dimensional, we choose an additional low-dimensional embedding $\gamma \in \Gamma$ to map to $\Theta$ with mapping function $\eta$ such that $\theta = \eta(\gamma)$.

Now, analogous to VAE, we model $q_{\phi_\gamma}(\gamma | \boldsymbol{x})$ as a Gaussian distribution and learn its parameters, i.e., $(\boldsymbol{\mu}_{\gamma}, \Sigma_\gamma)$, by training an encoder $E_{\gamma}$. The encoder, $E_{\gamma}$ essentially learns to map an observed data $\boldsymbol{x}=(\boldsymbol{x}_0,\ldots,\boldsymbol{x}_{T-1})$ (at all $T$ time points) to $(\boldsymbol{\mu}_{\gamma}, \Sigma_\gamma)$:
\vspace{-4pt}
\begin{align}
    (\boldsymbol{\mu}_{\gamma}, \Sigma_{\gamma}) = E_{\gamma}(\boldsymbol{x}) %
    \label{eq:gamma_enc}
\end{align}
{\em Remark: We use continous activation functions in $f_\theta$. So, any instantiation of the NN is continuous. Peano existence theorem guarantees the existence of at least one solution to the above ODE locally.}

We have now collected the components to (a) sample from $q(\gamma| \boldsymbol{x})$ and (b) map the embedding $\gamma$ to the NN parameter $\theta$ by learning $\eta$. This gives us a sampler to sample $f_\theta$ to generate a solution of the corresponding Neural ODE. 
\paragraph{Estimate initial condition $\z_0$ with a Data Encoder $E_{\z_0}$:}
Similar to the process of learning the latent representation of the transition functions,
we learn the distribution of initial values $\z_0$ as $q_{\phi_{E}}(\z_0 | \boldsymbol{x})=\mathcal{N}(\boldsymbol{\mu}_{\z_0}, \Sigma_{\z_0})$, by training an encoder $E_{\z_0}$ to map observed data $\boldsymbol{x}=(\boldsymbol{x}_0,\ldots,\boldsymbol{x}_{T-1})$ (at all $T$ time points)  to 
the latent space of $\mathcal{Z}$ whose samples are denoted by $\z_t$ via parameters $(\boldsymbol{\mu}_{\z_0}, \Sigma_{\z_0})$:
\vspace{-10pt}
\begin{align}
    (\boldsymbol{\mu}_{\z_0}, \Sigma_{\z_0}) = E_{\z_0}(\boldsymbol{x}) %
    \label{eq:z0_enc}
\end{align}
Now, with the initial point, $\z_0$ and transition function $f(\z_t, t)$ in hand, the remaining step is to map $\z_t$ to the corresponding $\boldsymbol{x}_t$ in the original data space. 

\paragraph{Mapping $\z$ to $\boldsymbol{x}$ via decoder $D$:}
Given the initial value $\z_0$ and transition function $f(\z_t, t)$, we can use existing ODE solvers to compute $\z_t$.
Once we obtain $\z_t$ using a non-linear function $D$, we recover the corresponding output $\boldsymbol{x}_t$, at time $t$. We model the output of the process $\boldsymbol{x}_t$ as a non-linear transformation of the latent measure of progression $\z_t$ as:
\begin{equation}
\boldsymbol{x}_t = D(\z_t) + \boldsymbol{\epsilon}_t,
\label{eq:dec}
\end{equation}
where $\boldsymbol{\epsilon}_t$ is measurement error at each time point.
This idea has been variously used in the literature \citep{pierson2019inferring, hyun2016stgp, whitaker2017bayesian}.

\paragraph{The final model:}
Combining \eqref{eq:gamma_enc}, \eqref{eq:z0_enc}, and \eqref{eq:dec} we get our Functional ODE model in  \eqref{eq:fode}, see Fig.~\ref{fig:model}. 
\begin{equation}
\left[
\begin{array}{l}
\z_{0} \sim \mathcal{N}(\boldsymbol{\mu}_{\z_0}, \Sigma_{\z_0})\text{, where } \boldsymbol{\mu}_{\z_0}, \Sigma_{\z_0} = E_{\z_0}(\boldsymbol{x})\\
\gamma \sim \mathcal{N}(\boldsymbol{\mu}_{\gamma}, \Sigma_\gamma)\text{, where } \boldsymbol{\mu}_{\gamma}, \Sigma_\gamma = E_\gamma(\boldsymbol{x})\\
\theta = \eta(\gamma) \mapsto f_\theta\\
\dot{\z}_t=f_\theta\left(\z_t, t\right)\\
\boldsymbol{x}_t = D(\z_t) + \boldsymbol{\epsilon}_t
\end{array}
\right.
\label{eq:fode}
\end{equation}

{\em Summary.} In contrast to Neural ODE and its variants, for a data $\boldsymbol{x}$, we not only use the encoder $E_{z_0}$ to map the observed data $\boldsymbol{x}$ to parameters of the distribution of ODE initialization $\z_0$, but we also utilize $\boldsymbol{x}$ and $E_\gamma$ to sample from function (or embedding) space of NN, characterizing trajectories. 
Given samples $\z_0$ and $\gamma$ (and thus corresponding $\theta$ of $f_\theta(\z_t, t)$), the $\z_t$ are computed as a solution to corresponding ODE, and $D$ is used to map $\z_t$ to $\boldsymbol{x}_t$ in the original space. Fig.~\ref{fig:model} shows an overview.
We now show how we can derive ELBO-like likelihood bounds used to train the model in \eqref{eq:fode}.

\subsection{ELBO for Functional ODE} 

The training scheme seeks to learn \begin{inparaenum}[\bfseries (a)] \item a distribution of $\z_0$, \item a distribution of $\gamma$, which is used to get the corresponding $\theta$ by learning \item a mapping function $\eta$, represented by NN.\end{inparaenum} 
At a high level, our approach is to infer the subspace of the function space, samples from which describe the trajectories, $f_\theta$, of the observed temporal process $\left\{\boldsymbol{x}_t\right\}$. We want to learn two components: an underlying simpler distribution for $\gamma$, i.e., $q_{\phi_\gamma}(\gamma | \boldsymbol{x})$, and mapping $\eta$, which maps $\gamma$ to parameters $\theta$, in-turn defining a sample, $f_\theta$, from the function/embedding space of NNs. 
A common strategy to learn the underlying distributions, $q_{\phi_\gamma}(\gamma | \boldsymbol{x})$ and $q_{\phi_{E}}(\z_0 | \boldsymbol{x})$, is to use a VAE \citep{chen2016variational}. It is known that in such probabilistic models, one wants to optimize the likelihood of data $\boldsymbol{x}$, but computing this likelihood $p(\boldsymbol{x})$ is usually intractable. Thus, we use a lower bound as  described next. 

Let $p(\z_0, \gamma)$ be the joint prior distribution, $q(\z_0, \gamma)$ be an approximate joint posterior. Let $p(\boldsymbol{x} \mid \z_0, \gamma)$ be the likelihood of reconstruction, where $\boldsymbol{x}=(\boldsymbol{x}_0,\ldots, \boldsymbol{x}_{T-1})$ is a function of the latent representation $\left\{\z_t\right\}$. 
Note that we have two sources of randomness: initial value of the latent trajectory $\z_0$, and a latent representation, $\gamma$, of a sample from function space of NNs. 
Using the marginalization property of probabilities, we can write the log likelihood of the data as $\log p(\boldsymbol{x})=\log \int p(\boldsymbol{x}, \z_0, \gamma) d(\z_0, \gamma)$, here $d(.,.)$ is the measure for integral. 
Using concepts from \S\ref{sec:background}, we can derive a lower bound for the $p(\mathbf{x})$ of our Functional NODE model as:

\vspace*{-2em}
\begin{align}
\log p(\boldsymbol{x})&=%
\log \int p(\boldsymbol{x} | \z_0, \gamma) p(\z_0, \gamma) \frac{q(\z_0, \gamma)}{q(\z_0, \gamma)} d(\z_0, \gamma)\label{eq:marg_prop}\\
&=\log \mathbb{E}_{q(\z_0, \gamma)}\left(p(\boldsymbol{x} \mid \z_0, \gamma) \cdot \frac{p(\z_0, \gamma)}{q(\z_0, \gamma)}\right)\nonumber\\
\geq \mathbb{E}_{q(\z_0, \gamma)}&\log p\left(\boldsymbol{x} | \z_0, \gamma\right)-KL\left(q(\z_0, \gamma) \| p(\z_0, \gamma)\right),\label{eq:elbo_fode_prelim}
\end{align}
where  \eqref{eq:marg_prop} follows from the definition of joint distribution and using Jensen's inequality.

Next, we define $q(\z_0, \gamma)$ and $p(\z_0, \gamma)$, which is needed to optimize the above lower bound (ELBO).

\paragraph{Choice of $q(\z_0, \gamma)$:} Recall that we identify the trajectory as a product of initial value (distribution $q_{\phi_{E}}(\z_0 | \boldsymbol{x})$) and transition function (distribution $q_{\phi_\gamma}(\gamma | \boldsymbol{x})$). Thus, we want to find disentangled distribution of $\z_0$ and $\gamma$.  
This leads to a form where $q(\z_0, \gamma)=q_{\phi_{E}}(\z_0 | \boldsymbol{x}) q_{\phi_\gamma}(\gamma | \boldsymbol{x})$. 
Different choices are 
available to define $q_{\phi_{E}}(\z_0 | \boldsymbol{x})$ and $q_{\phi_\gamma}(\gamma | \boldsymbol{x})$ and priors $p(\z_0)$ and $p(\gamma)$, e.g., Horseshoe \citep{carvalho2009handling}, spike-and-slab with Laplacian spike \citep{deng2019adaptive} or Dirac spike \citep{bai2020efficient}; For our experiments, modeling $q_{\phi_{E}}(\z_0 | \boldsymbol{x})$, $q_{\phi_\gamma}(\gamma | \boldsymbol{x})$ and $p(\z_0)$, $p(\gamma)$ as a Gaussian distribution leads to 
good performance (reconstruction and sample diversity). 
Other distributions will nonetheless be useful 
for other real-world time-series datasets. 
\vspace{-8pt}

\paragraph{Training:}
With all pieces in hand, we summarize the training of functional NODE (FNODE),
\begin{mybox}{gray}{}
Given the observation of temporal process, $\boldsymbol{x}$: 
\begin{compactenum}
\item Use encoder $E_{\z_0}: \boldsymbol{x}=\{\boldsymbol{x}_t\} \mapsto q_0$ and further sample initial point, $\z_0$ from $q_0$.
\item Use encoder $E_\gamma: \boldsymbol{x}=\{\boldsymbol{x}_t\} \rightarrow q_\gamma$, and sample latent representation of transition function, $\gamma$ from $q_\gamma$. 
\item Use mapping $\eta$ to map $\gamma$ to define weights of the NN $f_\theta$ and hence characterize $f(\z_t, t)$.
\item Use decoder $D$, fit 
the FNODE model, by minimizing the loss in \eqref{eq:elbo_fode_prelim}
\end{compactenum}
The output from this phase 
is the latent representation of both components (a) initial value $\z_0$ and (b) transition function $f(\z_t, t)$, which allows us to sample temporal processes as in common VAE settings.
\end{mybox}

Note that if the trajectories are the same for all training samples, 
then $\eta$ might collapse to a degenerate distribution due to the singularity. 
To avoid collapse, we use KL weight annealing (from 0 to 1), \cite{huang2018improving}. 

{\bf Correcting posterior inaccuracies:}
\begin{figure}[t]
\centering
\includegraphics[width=\columnwidth]{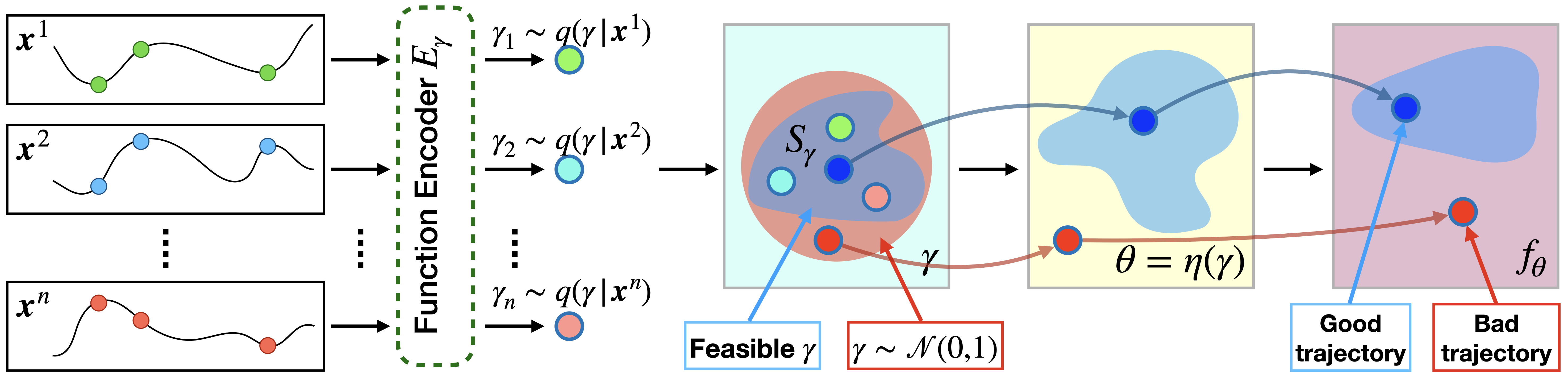} 
\caption{\footnotesize The direct sampling of $\gamma$ from $N(0, 1)$ results in a bad reconstruction of trajectories. However, given the samples from the learned approximate posterior distribution $q(\gamma|\boldsymbol{x}^i)$, we can fit a better distribution on top of these sample, refereed as $S_\gamma$, to generate proper trajectories. The fitting is happened after we train our model.
\label{fig:sampler}}
\end{figure}

While the variational setup is effective in deriving and optimizing for the maximum likelihood, prior results \citep{Kingma2016iaf} have observed that the Gaussian assumption on the prior $p(\z_0, \gamma)$ can often be too restrictive. This results in discrepancies between the prior $p(\z_0, \gamma)$ and marginalized approximate posterior $q(\z_0, \gamma)$.

To remedy this situation, using more flexible and sometimes hierarchical priors for VAE training \citep{Kingma2016iaf,Arash2020nvae} is useful. However, this leads to additional complexity for optimization since the computation of KL divergence $KL\left(q\left(\z_0, \gamma\right) | p\left(\z_0, \gamma\right)\right)$ can be non-trivial for non-Gaussian priors. Instead, inspired by \citet{Ghosh2020det}, we empirically found that fitting a Gaussian Mixture Model (GMM) on the learned embeddings $\z_0, \gamma$ in an ex-post manner accounts for the multimodal nature of the data and leads to drastic improvement in sample quality. Since the GMM is learned independently after VAE training on the low-dimensional embeddings, the additional computational complexity is negligible. We refer to this GMM model as $S_\gamma$ in the sequel.

\paragraph{Implementation:} For each sample $x^j$ from the training data, we encode parameters $(\mu_\gamma^j, \Sigma_\gamma^j)$ and draw $n_\gamma$ independent samples of $\gamma^j \sim N(\mu_\gamma^j, \Sigma_\gamma^j)$. Thus, given $N$ training samples, we obtain $N \times n_\gamma$ samples of $\gamma$, using which a GMM is learned. The motivation for fitting a GMM on samples of $N(\mu_\gamma^j, \Sigma_\gamma^j)$ instead of simply the means $\mu_\gamma^j$ is to also account for the variances $\Sigma_\gamma^j$ -- the target distribution to approximate is $q(\gamma)$ rather than $q(\mu_\gamma)$. More details on fitting the GMM and selecting the number of components is in the supplement.

\subsection{Statistical inference}
\label{sec:statinf}
Given a learned $S_\gamma$ we can perform different kinds of statistical inference. 
\paragraph{Sample generation:}
Given an initial value $\z_0$, obtained by encoding some $\boldsymbol{x}^i$ with $E_{z_0}$:

\begin{mybox}{gray}{}
\begin{compactenum}
\item To sample new trajectories, we can directly sample from $S_\gamma$.
\item To sample trajectories described by an example  $\boldsymbol{x}^j$, we can derive $\gamma_j$ corresponding to $\boldsymbol{x}^j$ through the encoder $E_\gamma$, and use it to generate a new trajectory
from the given $\z_0$, but which emulates the behavior of $\boldsymbol{x}^j$.
\item Relevant to example $\boldsymbol{x}^j$, we can also sample $\gamma\prime \in S_\gamma$, s.t. $|\gamma\prime - \gamma_j| \leq \delta$. Then, varying a value $\delta$, we can generate new trajectories closer or farther from example $\boldsymbol{x}^j$.
\end{compactenum}
\end{mybox}
\vspace{-5pt}
\paragraph{Uncertainty estimation:}
At test time for the given observed temporal process $\boldsymbol{x}^j$, we are
able to sample trajectories from the learned approximate posterior distribution 
$q_{\phi_\gamma}(\gamma | \boldsymbol{x}^j)$ and compute 95\% credible intervals (similar to confidence intervals but from a Bayesian perspective), posterior mean and variance. These summary statistics can express the uncertainty regarding temporal trajectories.

\paragraph{Out-of-Distribution (OOD) detection:}
OOD detection on trajectories is one potential application. In such a setting, outliers are defined by trajectories that have low-probability in the data-generating distribution, and thus unlikely to appear in the dataset. 

For conventional temporal models, performing OOD detection is particularly challenging because we must integrate the conditional probabilities $p(x_t| x_{[0, t)})$ over the entire time span $t \in [0, T]$. However, since our method treats trajectories simply as embedded functional instances on which a statistical model is learned directly, there is no need to integrate over the trajectory in order to perform likelihood estimation or statistical testing, which greatly simplifies the formulation.

\paragraph{Advantages:}
Assuming that the initial condition is fixed for a NODE, there is no variation in the trajectory -- so, there is no distribution to setup a statistical test. But in our case, statistical inference is still possible via sampling. 
Compared to BNNs,
via sampling $\gamma$, we use a valid likelihood test based on GMM fitted to sampled $\gamma$'s. Doing this with BNNs (assuming training succeeds) would either need comparisons of actual trajectories or a test on the posteriors of the entire network's weights.

\section{Related work}

{\bf Bayesian models.} While our formulation is different from Neural ODEs (where only the initial condition of the ODE varies), there are also distinct differences from Bayesian versions of NODE \citep{yildiz2019ode2vae}, where $f$ is parameterized by a Bayesian Neural Network. 
In general, Bayesian Neural Networks (BNN) \citep{goan2020bayesian} are not considered  generative models, like VAE or GAN \citep{goodfellow2014generative}.
While VAE-based models, as in our work, seek to learn a latent representation of the data (temporal process), the main goal in BNNs is to account for uncertainty in the data, learning a posterior distribution of a {\em network}, given all the observed data.
In contrast to BNN, our work learns an encoding of the temporal process $\boldsymbol{x}^i$ in its own latent representation $q_{\phi_\gamma}(\gamma | \boldsymbol{x}^i)$ from which different trajectories are sampled. 
Conceptually, for each temporal process $\boldsymbol{x}^i$, the sampling from latent representations $\gamma_i$ with distribution $q_{\phi_\gamma}(\gamma | \boldsymbol{x}^i)$ can be mapped to a BNN form of NODE. However, since the learned latent representation can be considered as a mixture of models, our approach can be thought as a mixture of BNN ODEs. 
In addition, the latent representation of the trajectory $\gamma$ allows us to easily perform a likelihood-based test for Out of distribution detection, which is not directly available for Bayesian Network.

{\bf Neural ODEs/CDEs.} 
The reader will notice that we sample the entire neural network $f_\theta$, which is otherwise fixed in NODE \cite{chen2018neural} and hence each sample from the space of FNODE is a NODE. 
This argument can be made more precise. Consider the formulation given in \cite{kidger2020neural}. That result (theorem statement in \S3.3) notes that any expression of the form $z_t = z_0 + \int_{t_0}^t h_\theta (z_s, X_s) ds$ may be represented exactly by a Neural controlled differential equation (NCDE) of the form as $z_t = z_0 + \int_{t_0}^t f_\theta (z_s) dX_s$. But the converse statement 
does not hold. In our case, the different $\gamma$’s precisely correspond to different $X$’s i.e., they are functions controlling the evolution of the trajectory. More specifically, $\gamma_1$ is the parameterization for a control path (function) $X_1$, $\gamma_2$ for $X_2$ and so on. Note that these functions are evaluated along the domain of definition as usual, meaning that $X_1(s) \equiv X_{1_s}$ is the value of the function $X_1$ at the point $s$. Therefore, as in \cite{kidger2020neural}, our model generalizes NODE in a similar way as NCDE. The distinction between NCDE and FNODE is that the latter deals with modeling a distribution over the control paths (functions) rather than simply one path as in \cite{kidger2020neural}.

\begin{figure*}[!t]
    \centering

  \includegraphics[width=0.39\textwidth]{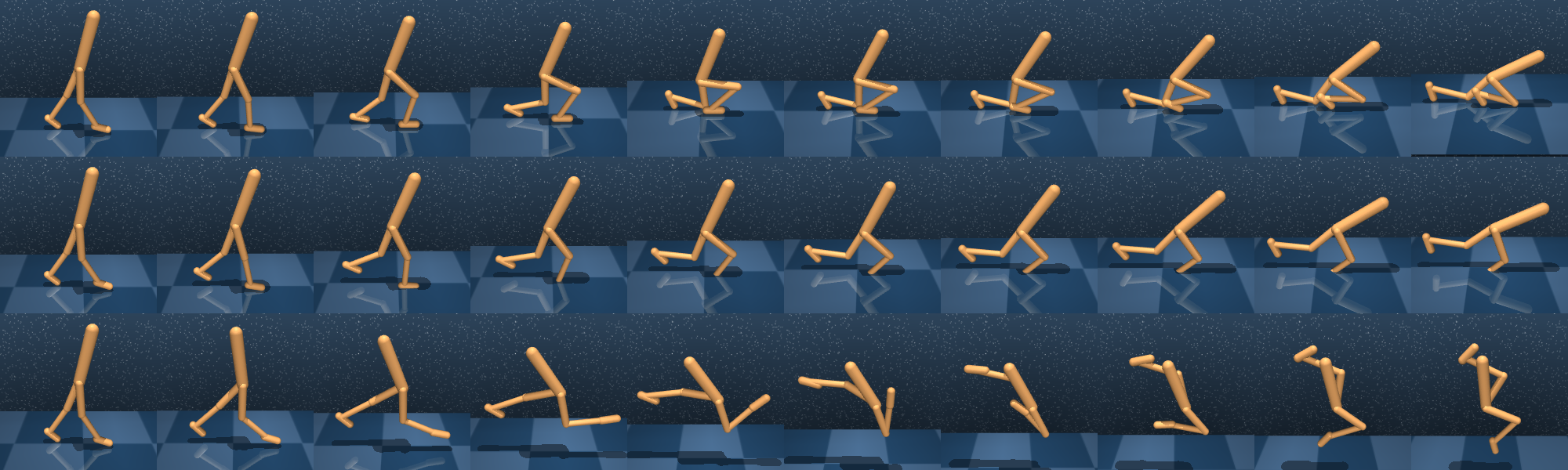}
  \includegraphics[width=0.39\textwidth]{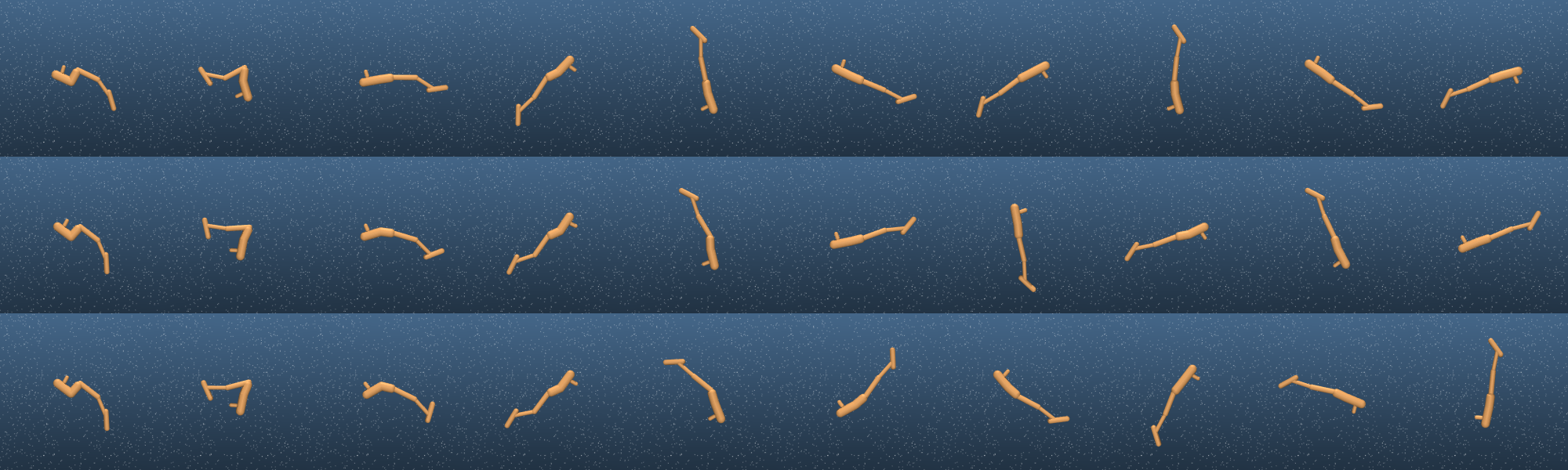}
  \includegraphics[width=0.198\textwidth, trim={0cm, 0cm, 35cm, 0cm}, clip]{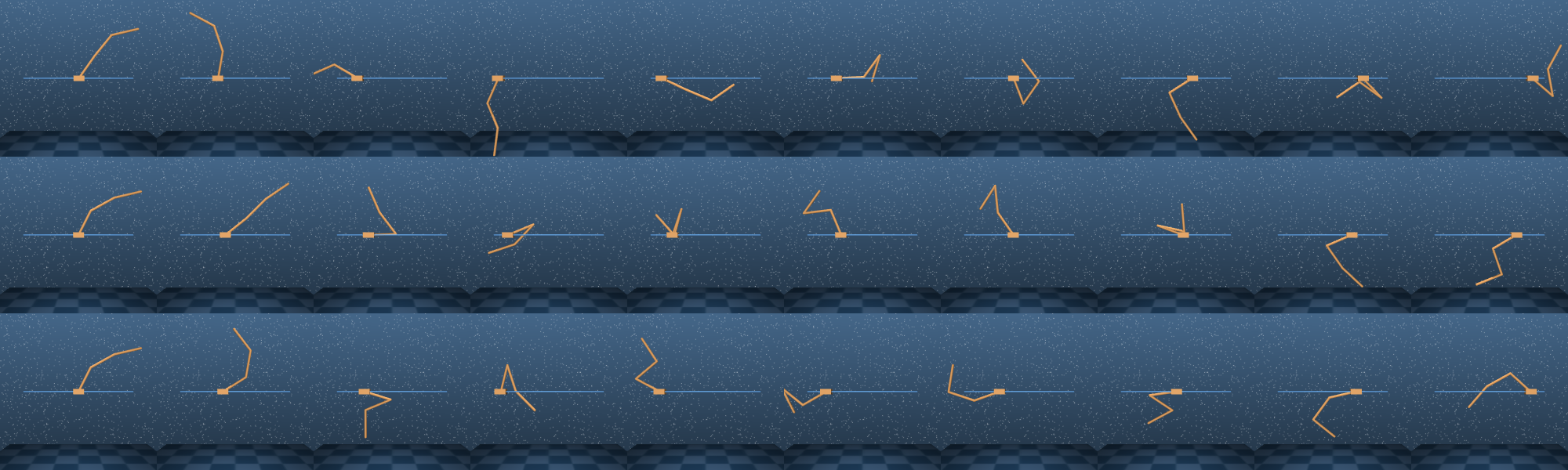}

  \includegraphics[width=0.39\textwidth]{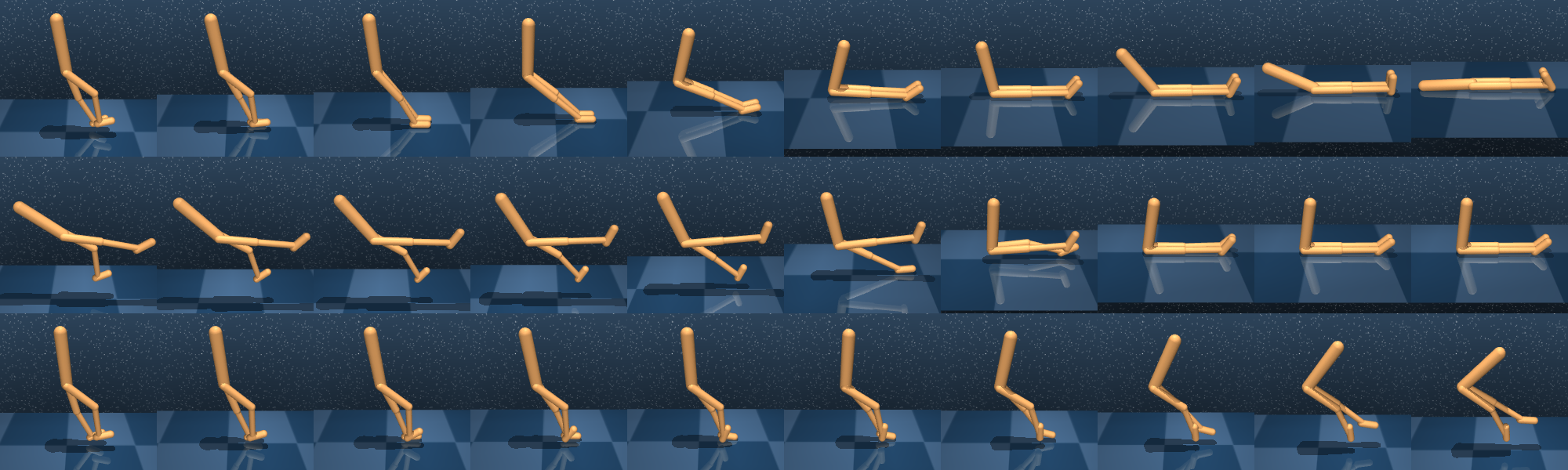}
  \includegraphics[width=0.39\textwidth]{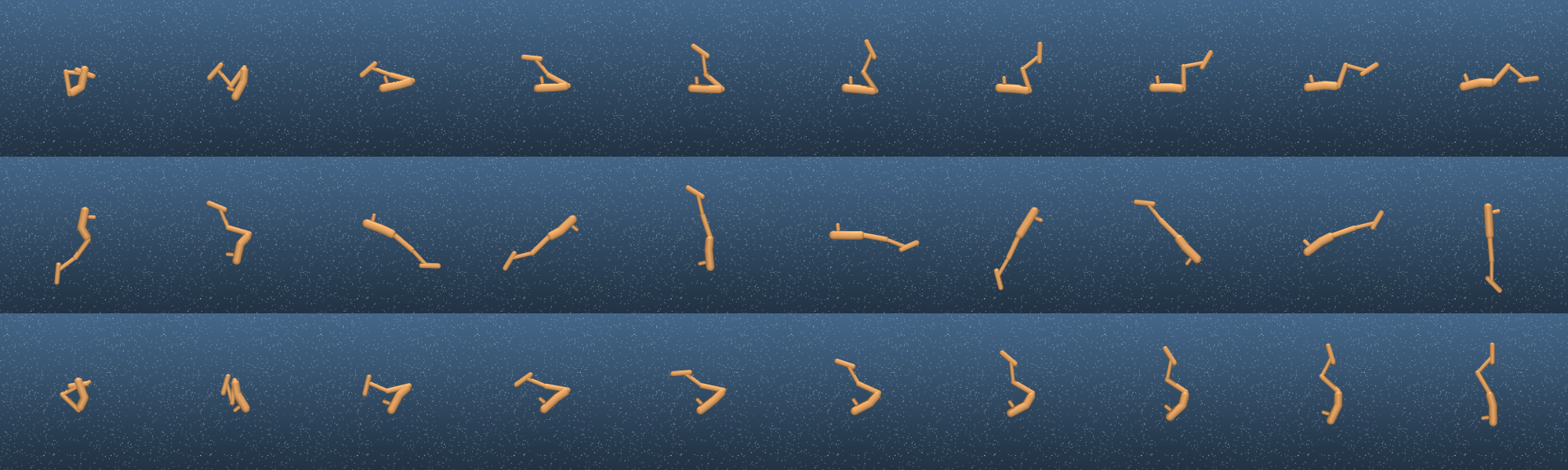}
  \includegraphics[width=0.198\textwidth, trim={0cm, 0cm, 35cm, 0cm}, clip]{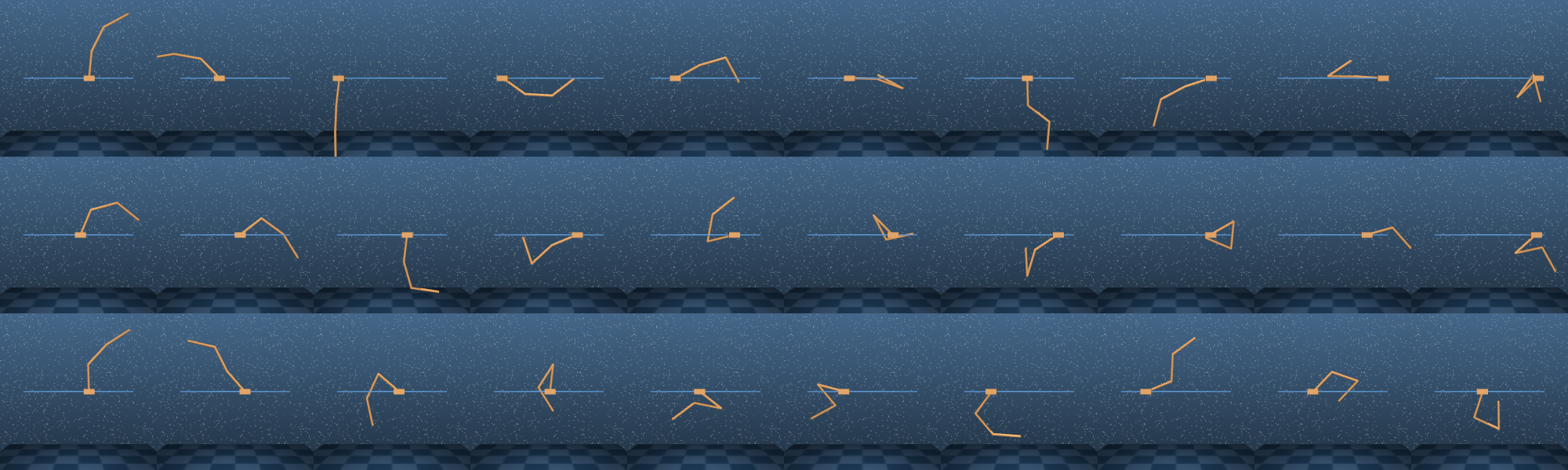}

\caption{\footnotesize We present visualization of generated trajectories of three different data sets from MuJoCo. From left to right: Walker, Hopper, and 3 poles cartpole. \textbf{Top row:} Trajectories are generated by fixing initial condition $\z_0$ and sampling $\gamma \sim S_\gamma$, \textbf{Bottom row:} We transfer trajectory by sampling $\gamma$ from the exemplar, and applying it to $\z_0$. First row in each block represent original data from which we derive $\z_0$, second row is an exemplar used to transfer trajectory, and last is sampled/transferred.}
\label{fig:mujoco}
\end{figure*}

\section{Experiments}
We evaluate our model on various temporal datasets, of varying difficulty in the underlying trajectories, representing physical processes, vision data, and human activity. Namely: \begin{inparaenum}[\bfseries (1)] 
\item 3 different datasets from MuJoCo,
\item rotating/moving MNIST, and
\item a longer (100 time steps) real-life data, describing human body pose, NTU-RGBD.
\end{inparaenum}
Other experiments are in the supplement.

{\bf Goals.} 
We evaluate: \begin{inparaenum}[\bfseries (a)] 
\item  ability to learn latent representations of the samples of function space, i.e., representation of NN, 
\item ability to generate new trajectories, including the trajectories based on examples
\item the ability to perform an Out-of-Distribution analysis for the trajectory and other kinds of statistical inference.
\end{inparaenum}
We defer details of hardware, neural network architectures, including encoders/decoder and hyperparameters to the supplement.

{\bf Baselines.} We compare FNODE with several probabilistic temporal models, namely: 

{\bf(a)} VAE-RNN \cite{chung2015recurrent}, where the full trajectory is encoded in the latent representation;
{\bf(b)} Neural ODE (NODE) \cite{rubanova2019latent}, where only initial point of ODE is considered to be stochastic;
{\bf(c)} Bayesian Neural ODE (BNN-NODE)\cite{yildiz2019ode2vae}, where the stochasticity is modeled through initial value $\z_0$ and representing transition function $f$, using Bayesian version of Neural ODE;  
{\bf(d)} Mixed-Effect Neural ODE \cite{nazarovs2021variational}, where both components, initial value $\z_0$ and transition function ff, through mixed-effects.

\subsection{Variational Sampling of trajectories}
We start by showing FNODE's ability to sample trajectories. Mainly we are interested in validating: 
(a) for the same initial condition, can our model generate wide variety of trajectories?
(b) can our model ``transfer" trajectory of an example $\boldsymbol{x}^j$ to another temporal process $\boldsymbol{x}^i$?

{\bf MuJoCo:}
Data from MuJoCo \citep{todorov2012mujoco} provides accurate simulation of articulated structures interacting with their environment to represents simple Newtonian physics. We vary the difficulty of the temporal process, by considering different objects and tasks: 
humanoid walker,
humanoid hopper, and 
3 poles cartpole which is a moving along the x-axis cart with rotation of 3 joints, attached to the base.
{\em Sample trajectories:} For each experiment, we randomly generated initial conditions and applied 10 different forces, then we observed 20 time steps.  
To make visualization easier, we plot every other step.
In Figure \ref{fig:mujoco}(top), we show trajectories, generated from the same initial value $\z_0$, but varying samples of the $\gamma$ (latent representation of the transition function). For all three datasets, we are able to generate diverse and in-distribution trajectories. 
{\em Transfer trajectory to other temporal process:} We present a few examples  of transferring trajectories in Figure\ref{fig:mujoco}(Bottom).

\begin{figure}
\centering 
  \includegraphics[width=0.44\textwidth]{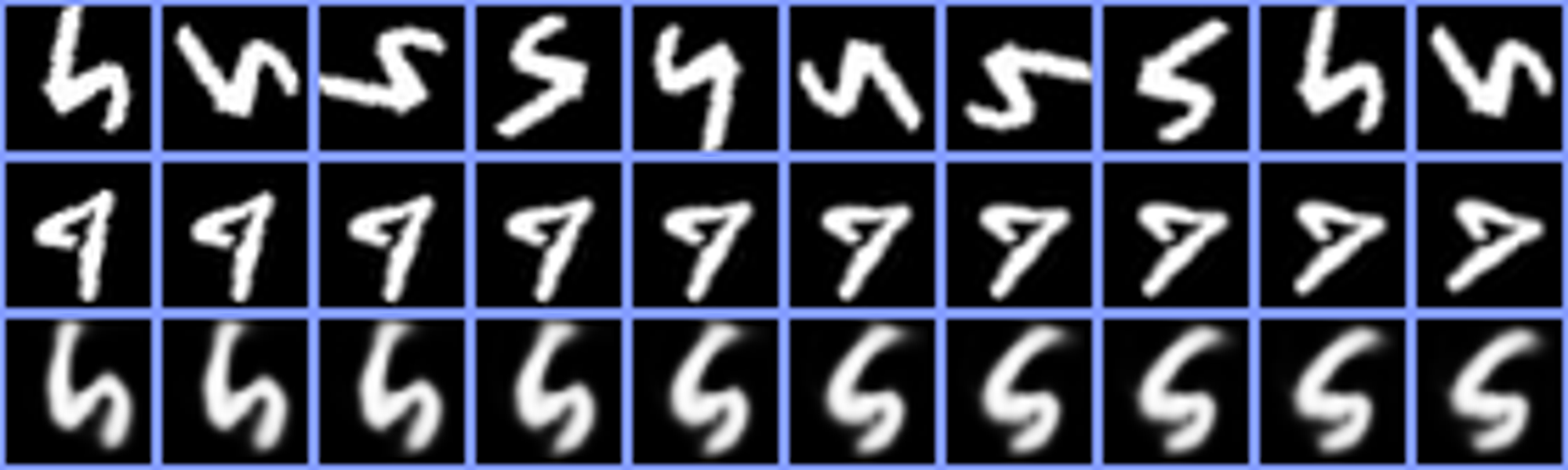}
  \includegraphics[width=0.44\textwidth]{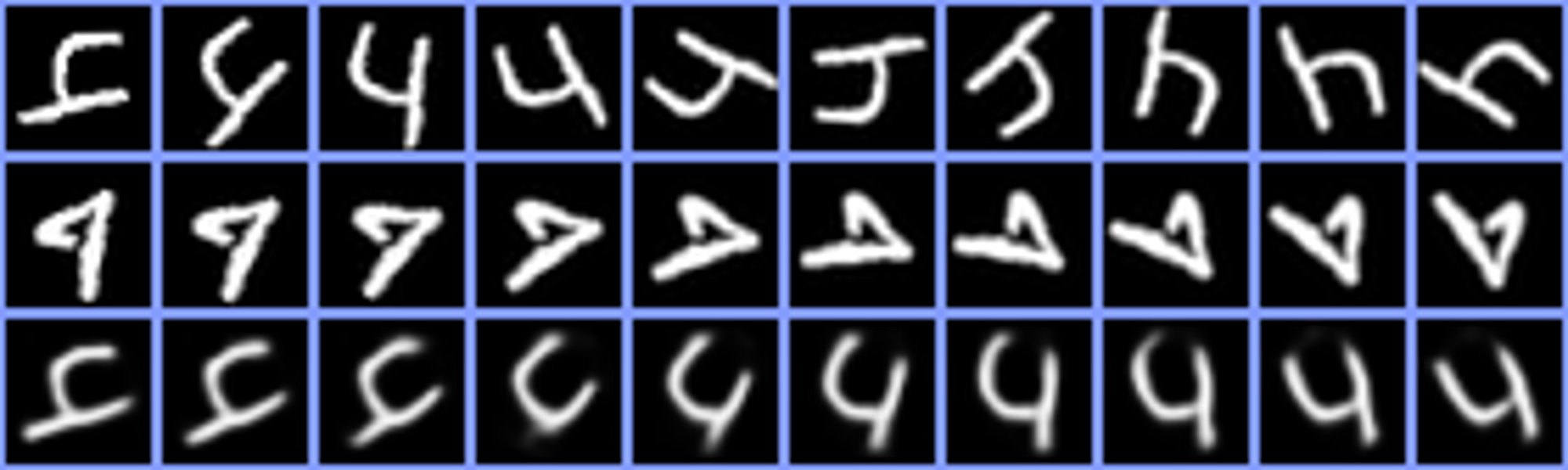}
\caption{\footnotesize Row $1$ is a process used to derive $\z_0$, row $2$ is an exemplar, projection of which we would like to transfer, and third row is a result of transfer. Notice how the speed of row $2$ transfers to the first row, given an initial condition.}
\label{fig:rotmnist}
\end{figure}
In Fig.\ref{fig:mujoco}, comparing the temporal process in row $1$ (from which we derive $\z_0$) and the example in the row $2$ (from which we derive $\gamma$), there is a large difference in the speed of walker. In row $2$ the example needs almost the entire row to fall down from a small height, while the original fall (first row) occurs at a faster pace and from a higher height. As expected, after transfer, the resulting trajectory of our walker (row $4$) does not fall down completely, because the pace is slower, derived from row $2$, while initial position is at a similar height as in row $1$. Similar observations can be made for hopper (2nd set): different speeds of rotation/unbending; and for cartpole: different directions of xx-axis movement (3rd set).

{\bf Rotating/Moving MNIST:}
While MuJoCo provides a wide variety of trajectories, it pertains to the same object. Here, we want to check whether the sampling of trajectories based on an example, works not only within the same class of digits (similar to MuJoCo), but across different styles of Rotating and Moving MNIST.

\begin{figure}
\begin{subfigure}{0.49\columnwidth}
\includegraphics[width=\linewidth]{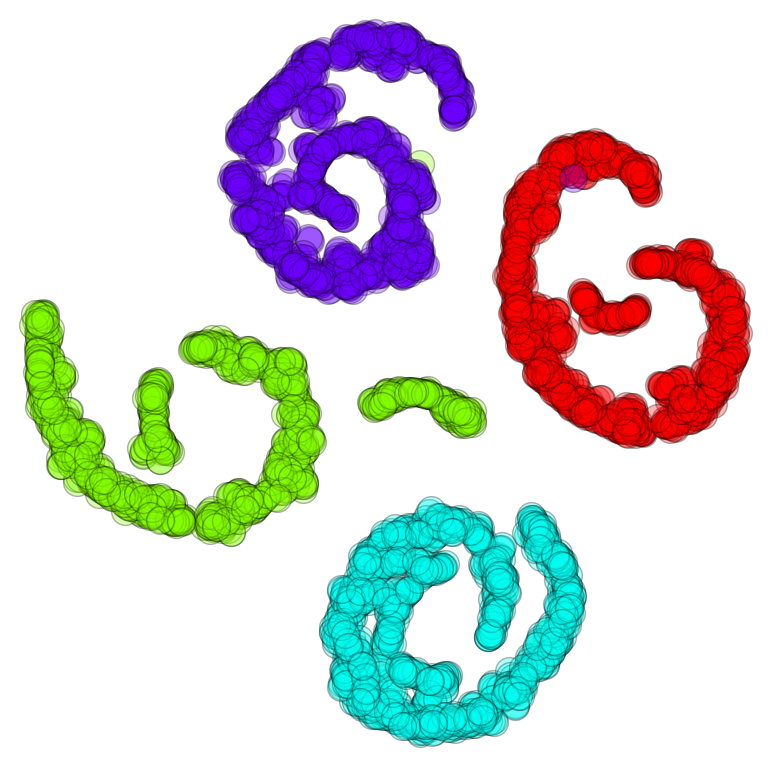}
\caption{\footnotesize  {TSNE representation of $\gamma$. Colors correspond to the angle of rotation.}}
\label{fig:mnist_tsne}
\end{subfigure}
\begin{subfigure}{0.49\columnwidth}
\includegraphics[width=\linewidth]{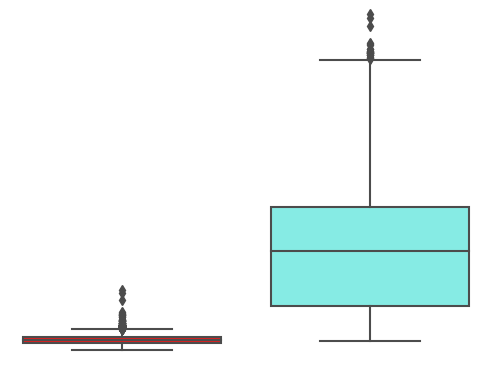}
\caption{\footnotesize Moving MNIST OOD: red (set $A$), blue (set $B$). Boxplots of negative log-likelihood. Negative log-likelihood of data for the set B is much higher.}
\label{fig:movingmnist_tsne}
\end{subfigure}
\end{figure}
We model Rotating MNIST by 3 random components: digit (from 0 to 9, any style), random initial rotation, and 10 different angles of rotation varying from $-45^{\circ}$ to $45^{\circ}$. 
For training, we used data observed for $10$ time steps. 
For Moving MNIST we generate 4 random components: digit (from 0 to 9, any style), random initial position and velocity, and 5 different speeds of movement varying from $5$ to $50$. 

\begin{table*}[tb]
\centering
\resizebox{1\textwidth}{!}
{%
\begin{tabular}{c|*{5}{c}|*{5}{c}}
\specialrule{1pt}{1pt}{0pt}
 & \multicolumn{10}{c}{Data}\\
\cline{2-11}
& \multicolumn{5}{c|}{Rotating MNIST} & \multicolumn{5}{c}{Moving MNIST}\\
\cline{2-11}
 &  & \multicolumn{4}{c|}{Extrapolation}  & & \multicolumn{4}{c}{Extrapolation}  \\
\multirow{-4}{*}{Model} & \multirow{-2}{*}{Interpolation}  & 10\% & 20\% & 50\% & 100\% & \multirow{-2}{*}{Interpolation}  & 10\% & 20\% & 50\% & 100\% \\
\hline
 VAE-RNN  &0.0206&0.0865&0.0850&0.0840&0.0831  &0.0120&0.0565&0.0587&0.0578&0.0579   \\
 NODE & 0.0169 &0.0903&0.0904&0.0887&0.0863  &0.0161&0.0534&0.0566&0.0553&0.0554   \\
 BNN-NODE &0.0193&0.0870&0.0870&0.0856&0.0834 &0.0208&0.0518&0.0555&0.0537&0.0534  \\
 ME-ODE  &0.0152&0.0877&0.0875&0.0870&0.0859  &0.0193&0.0522&0.0554&0.0552&0.0553   \\
 FNODE (Ours)  &0.0108&0.0929&0.0932&0.0925&0.0911  &0.0100&0.0554&0.0591&0.0590&0.0594   \\

\bottomrule
\end{tabular}
}
\caption{Interpolation and Extrapolation MSE, where extrapolation is computed in respect of \% of observed steps. Since all methods are probabilistic models, to evaluate MSE we compute the average among generated samples. While interpolation error of our method is lower compared to baselines, the extrapolation error is slightly higher. This happens because our method provides a wide variety of trajectories, so a higher variance.}
\vspace{0pt}
\label{tab:baselines}
\end{table*}

\begin{figure*}[bt]
    \centering 
    \begin{subfigure}{0.49\textwidth}
    \includegraphics[width=1\textwidth]{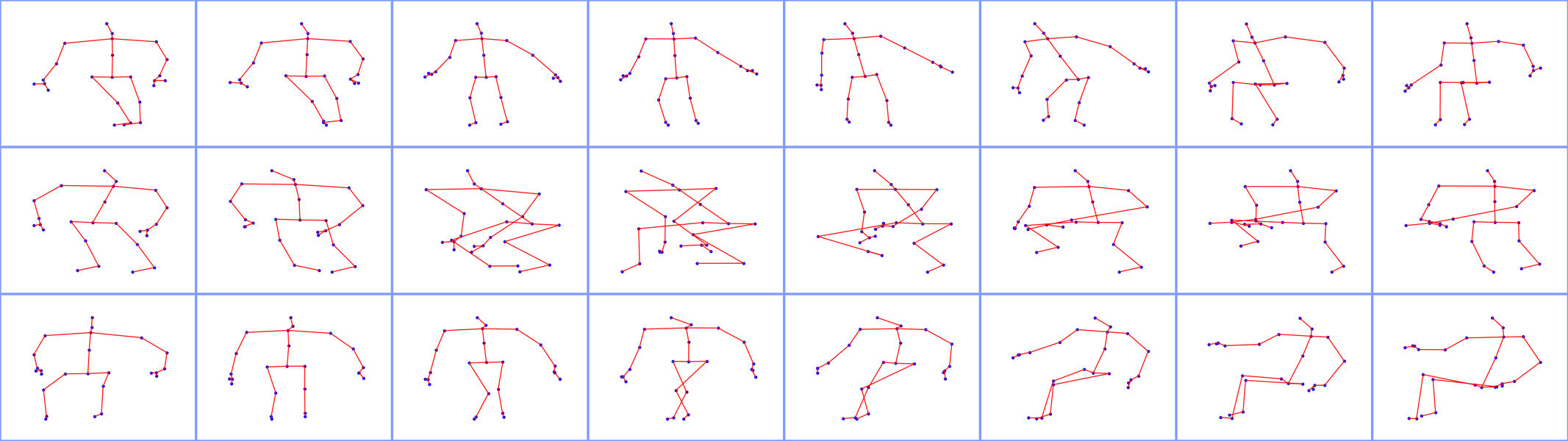}
    \caption{\footnotesize Initial trajectory and exemplar are actions of sitting,
    however, on the second row sitting is faster, which results in a fast sitting in the third row}
    \end{subfigure}
    \rulesep
    \begin{subfigure}{0.49\textwidth}
    \includegraphics[width=1\textwidth]{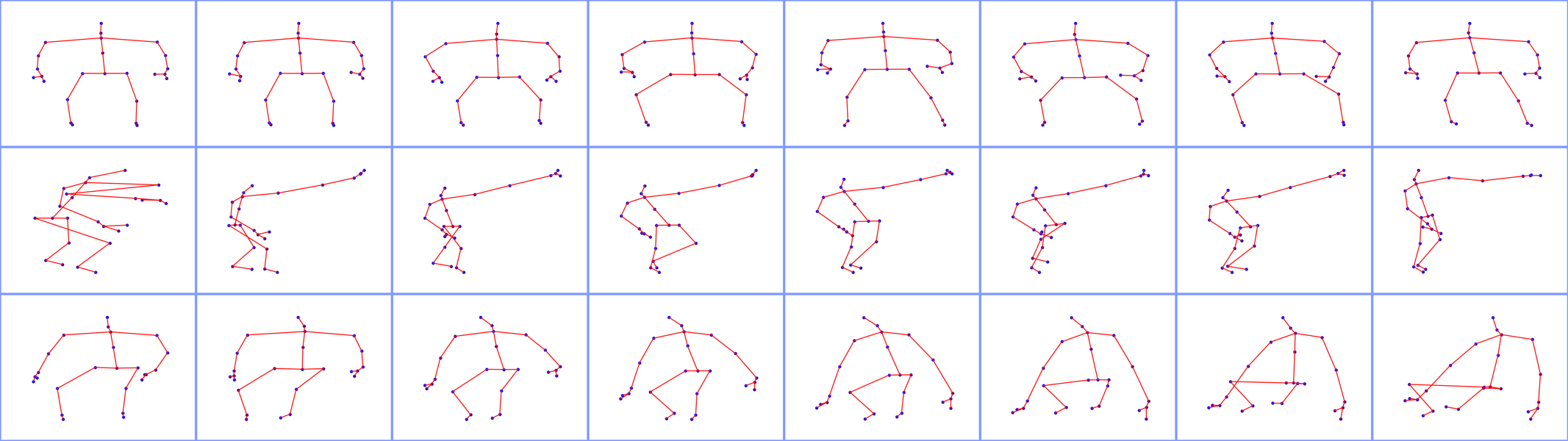}
    \caption{\footnotesize Initial trajectory is a sitting action, while exemplar is hand waving. This results in a novel motion, that in the third row, the person sits but with extended hands.}
    \label{fig:ntu_2}
    \end{subfigure}
\caption{\footnotesize Represent two different samples of transferring trajectory. First row is a process used to derive $\z_0$, second row is an exemplar, projection of which we would like to transfer, and third row is a result of transfer. Notice how with the exemplar we can `teach' another object to do similar things.}
\label{fig:ntu}
\end{figure*}

While our goal is variational sampling (and not extrapolating to predict future steps), FNODE provides sensible results for interpolation and extrapolation, compared to other baselines.
To this end, for both datasets we generate 20 time-steps. The first 10 time-steps are used as observed to fit the model and learn the distribution of $\z_0$ and $\gamma$. The other steps, from 11 to 20, are used to evaluate the ability of our model to extrapolate. 
In Tab.~\ref{tab:baselines}, we provide Mean Squared Error (MSE) for both interpolation and extrapolation. Notice that our method (FNODE) has a smaller interpolation error, compare to other baselines; however, the extrapolation error is higher. This is because we sample both initial value $\z_0$ and representations of the transfer function $\gamma$, so we have a wider range of trajectories. We provide plots in supplement.
Fig.~\ref{fig:rotmnist} shows the ability of our model to transfer the trajectory of an exemplar (second row) on the data from which we derived the initial value $\z_0$ (first row). The resultant transferred trajectory is shown in the third row. We see that our model can transfer trajectories across objects of different classes. This implicitly follows from the choice of our model of disentangling trajectory in 2 spaces: space of initial values $\z_0$ and parameterization of the transfer function $\gamma$. Notice how smoothly the speed of rotation is transferred to the original data generating a new trajectory. In Figure~\ref{fig:mnist_tsne}, we show a TSNE visualization of samples from $\gamma$, embedding of transition functions that parameterize trajectories.

{\bf Compute/memory efficiency:}
For batch size $8$ (rotation MNIST),  %
our {\bf (1)} memory need is $3\times$ more than VAE-RNN/NODE/ME-ODE, and $2\times$ more than BNN-NODE.
However, for {\bf (2)} wall-clock time, we need 2$\times$ compared to VAE-RNN/NODE, but only 0.75$\times$ of ME-ODE/BNN-NODE. 
Given the benefits of our model, it is a sensible trade-off. 

{\bf NTU:}
NTU  is a dataset for human actions recognition \citep{shahroudy2016ntu}. It consists of 126 states with varying number of time-steps. Each state is described by 26 joints in 3D coordinates, which are captured using Kinect v2.
We select data from $10$ states, spanning $100+$ time steps. To deal with the longer time series data in the NTU data, 
we use signature transforms \citep{morrill2021neural}.
We compute signature transform on the segments of $10$ steps with signature depth of $5$ (see \citep{morrill2021neural}). 
We show that despite using the summaries, we are still able to perform satisfactory generation of new trajectories in the observed space.
In Fig.~\ref{fig:ntu}, we show a 2D projection of human activity, generated by our model. 
Fig.~\ref{fig:ntu_2} shows an interesting trajectory, where the model learns from the exemplar to extend the hand, and incorporates this with the sitting action.

\subsection{Statistical inference}

We show that given a learned distribution of $\gamma$, we are able to perform an out-of-distribution analysis, to detect trajectories unfamiliar (or anomalous) for our learned model.

{\bf Moving MNIST:}
We create a dataset $A$ by sampling different MNIST digits, random initial position in the frame and direction and traveling at a given speed.
We train our model to learn the embedding of the transition function, $\gamma$. 
Then, we generate additional data traveling at a different velocity (call it $B$), resulting in a different set of trajectories. 
We conduct OOD analysis, following the description in Sec.\ref{sec:statinf}.
In Fig.~\ref{fig:movingmnist_tsne} boxplots of negative log-likelilhood  show a significant difference between groups. Clearly, implicit representation of trajectories using our method is beneficial for OOD.

{\bf NTU:}
We carry out a similar experiment on the NTU dataset.
However, here, we increase the number of unknown states in the set $B$.
The observed set $A$, used to train the model, contains information on only about 6 randomly selected classes of human activity. For the testing set $B$, we use the trajectories from 53 unobserved classes and perform the OOD testing as before. Due to a large number of states, instead of plotting 59 boxplots for negative log-likelihoods, we provide a table with the ratio of points per class, which is defined as an outlier for set $A$, Table~\ref{tab:ntu_ood}. We can see that almost 100\% of samples from a set $B$, not in the training set $A$, are identified as outliers.

\begin{table}[!tb]
\centering
\resizebox{\columnwidth}{!}
{\small
\begin{tabular}{l|ccccccc|ccccccc}
\specialrule{1pt}{1pt}{0pt}
\rowcolor{tableheadcolor}
\multicolumn{1}{c|}{\cellcolor{white}} & \multicolumn{7}{c|}{Set $A$} & \multicolumn{7}{c}{Set $B$} \\
\hline
\multicolumn{1}{c|}{Classes of Human Actions} & 
6 & 8 & 9 & 23 & 27 & 36 &  43 & \multicolumn{7}{c}{all} \\
\hline
     \multicolumn{1}{c|}{Proportion of outliers} &
      0.06 & 0.07 & 0.06 & 0.03 & 0.05 & 0.02 & 0.06 & 
      \multicolumn{7}{c}{$> 0.97$}\\
 
\bottomrule
\end{tabular}
}
\caption{ Proportion detected as outliers per class in: train set $A$ (in-distribution) and test set $B$ (OOD).}

\label{tab:ntu_ood}
\end{table}

\section{Conclusions}
We studied mechanisms enabling
generation of new trajectories of temporal processes based on variational sampling. 
Experiments on several real/synthetic datasets show that the sampling procedure is effective
in sampling new trajectories and in other statistical inference tasks such as OOD detection, topics that have remained underexplored in the generative modeling literature for temporal data. We note that very high dimensional sequences with complicated temporal dynamics remain out of reach at this time.

\clearpage

\appendix
\clearpage
\newpage

\vspace{-10pt}
\section{Limitations}
Several challenges or limitations in our formulation should be noted. First, since we parameterize $f_\theta$ through the computation of both $\gamma$ and $\eta(\gamma)$, it results in a larger computation graph for backpropagation than a standard NN with parameters as weights. This limitation requires a more careful architectural choice for $f_\theta$. 
Further, although using a Gaussian Mixture Model $S_\gamma$ significantly improves the performance of the data decoder to reconstruct unseen trajectories, yet we encounter cases where the performance is not comparable to those from a 
VAE/GAN for a single frame.  

\section{Societal impacts:}
Efficient probabilistic generative models for temporal trajectories, which can be used for uncertainty estimation and other statistical inference tasks such as out of distribution testing, is by and large, a positive development from the standpoint of trustworthy AI models.

\section{Hardware specifications and architecture of networks}
All experiments were executed on NVIDIA - 2080ti, and detailed code will be provided in the github repository later (currently it is attached).

\paragraph{Neural Differential Equations:} In any of the implementation we used Runge-Kutta 4 method with ode step size 0.1.
To model the equation $\dot{\z}_t=f(\z_t, t)$, we consider latent representation $\z_t$ to be in 1 dimension with $p$ channels. 
Recall that we learn initial value $\z_0$ from the data directly, and thus we are required to apply encoder.

\paragraph{Encoder $E_{\z_0}$: } For the $\z_0$ encoder we use a known and established ode-rnn encoder \cite{rubanova2019latent}. For the datasets, like Moving/Rotation MNIST and NTU in addition to ode-rnn encoder we apply a pre-encoder. The main purpose of pre-encoder is to map input in 1 dimensional data with $k$ channels. Which is later used by $E_{\z_0}$ to map into $p$ channels. 

For \textbf{Moving/Rotation MNIST} we use ResNet-18 pre-encoder, with following number of channels (8, 16, 32, 64) for an Encoder. 

For \textbf{NTU}, the used data  consist of 100 time steps with 26 different sensors, and each sesor is with 3d coordinates. As we mentioned in the main paper, we apply the signature transforms.
Given the sample with 100 time steps, we split it in chunks of 10, and apply signature transform of depth 5 for each chunk. This results in 10 chunks total with 363 channels (formula: $3 + 3^2 +\ldots + 3^5$).
As a result of signature transformation we obtain 10 time steps for 86 sensors and 363 channels each. We then apply several FC layers to change number of channels, then we collapse 26 sensors with resulted channels and convert it to $k$ channels, as mentioned above.
\begin{verbatim}
nn.Sequential(
    nn.Linear(363, 128), nn.ReLU(),
    nn.Linear(128, 256), nn.ReLU(),
    nn.Flatten(2),
    nn.Linear(last_dim * 256, 128), 
    nn.ReLU(),
    nn.Linear(128, output_dim))
\end{verbatim}

\paragraph{Encoder $E_\gamma$:} This is an important part of our modelling, which given temporal observations with $T$ time steps, $\boldsymbol{x}=(\boldsymbol{x}_1,\ldots,\boldsymbol{x}_T)$, encodes the information about trajectory into latent representation $\gamma$. For this we tried a lot of different encoders, including: \textbf{(a)} fully connected, \textbf{(b)}  neural-ode,  \textbf{(c)} dilated CNN-1d, \textbf{(d)}  LSTM,  \textbf{(e)} RNN, and \textbf{(f)} Time Series Transformer. To our surprise the fully connected network performed the best in respect of learning trajectory and training time. The second place was Time Series Transformer.

\paragraph{Projection $\eta: \gamma \implies \theta$:}
For the projection network $\eta$ we use a fully connected network, with Tanh() activation function between layers, and after the last layer. Because $\eta$ eventually returns us weights of a neural network $\theta$, which cannot be extremely big, it was necessary for us not only apply Tanh() as a last layer, but also introduce another single-value parameter $\lambda_\theta$, which is used to scale $\theta = \lambda_\theta \tanh(F(\gamma))$. We noticed a significant boost in performance once we introduced a $\tanh$ on the last layer and a learned scalar $\lambda_\theta$, compare to weights being completely unbounded.

\paragraph{NN parameterization of transition function $f$:} The previously described weights $\theta$, are the weights of NN, which parameterize $f$ in $\dot{\z}_t=f(\z_t, t)$. For all experiments we used $f$ as a fully connected network with 3 layers with 100 hidden units each layer and $\tanh$ activation function between layers. While the size of this network might look small, empirically it has enough power to model the changes in latent space over time.

\paragraph{Computation aspects of optimization:} (This paragraph represents an implementation in the PyTorch, however, it is similar for other popular choices like Tensorflow.) 
Since we model weight of the NN $f$, rather than consider them as parameters, we cannot use a standard implementation of layers in PyTorch, like `nn.Linear()'. Instead, we create a class of functional networks, where each layer is presented by an operation, like nn.functional.linear. The class is initialized by copying the structure of the provided network, but replacing layers with nn.functional correspondence. 
The forward function of the class accepts input $x$ and weights $\theta$. Then $\theta$ is sliced to fit the size of the current layer and layers is applied to $x$. 
As we mentioned in the Limitations of the main paper, this results in a larger than usual computation graph, and requires a more careful architectural choice for $f$.

\paragraph{Decoder $D_{\boldsymbol{x}}$:} 
For the Moving/Rotation MNIST,  similar to Encoder $E_{\z_0}$, we use ResNet-18 decoder, but with (32, 16, 8, 8)  channels. For other data sets we mainly use fully connected neural networks.

\section{Selecting GMM}
One of the components of our work is to utilize Gaussian Mixture Model (GMM) on top of samples $\gamma$ from the learned posterior distributions $q_{\phi_\gamma}(\gamma | \boldsymbol{x}^i)$. 
The architecture of GMM is based on 2 key factors: number of components and covariance type.
To select the best GMM model, which is later used for sampling from posterior distribution, we range number of components from 10 to 200, with a step 10, and consider 4 different types of covariance matrix provided in python machine learning package ``sklearn": ``spherical", ``tied", ``diag", ``full". Among all these models we select the one with lowest BIC score as the best.

\section{Experiments}
We start the experiment section showcasing a comprehensive analysis of a synthetic panel data set, which conceptually represents observations of multiple individuals over time. Our aim is to showcase the effectiveness of our framework in learning meaningful representations of the transition function $f$ of temporal trajectories.

\subsection{Complete analysis of a synthetic panel data}
To demonstrate the ability of our framework to learn meaningful representation of the transition function $f$ of temporal trajectories, we consider a synthetic panel data, which is represented in the form: 

\begin{equation}
\left[
\begin{array}{l}
x^{(i)}_t = A^{(i)} \cdot \sin(\psi^{(i)}_t) +\varepsilon^{(i)}\\
\frac{d\psi^{(i)}_t}{dt} = 2\pi\\
\varepsilon^{(i)}\sim N(0,10^{-3})
\end{array}
\right.
\label{eq:ch5_synthe_exp}
\end{equation}

In this representation, $x^{(i)}_t$ conceptually might represent the $i$ patient's observation at time $t$, for example, it can describe amyloid accumulation in the region of the brain \cite{betthauser2022multi}. The derivative of $x^{(i)}_t$ with respect to $t$ represents the rate of change of $x^{(i)}_t$ with respect to time. The amplitude of changes $A^{(i)}$ is defined per group $i$, e.g. patient, and can vary among groups. While $x^{(i)}_t$ can be a multi-dimensional vector, withot loss of generality we consider $x^{(i)}_t$ to be a one dimensional. Namely, for each group $i$ at time step $t$, $x^{(i)}_t$ is 1 dimensional vector.

For this set of experiments, we generate $10,000$ examples with $10$ unique values of $A^{(i)}\sim Unif(0, 10)$. In addition to make data more complicated in separation, and check that our framework is capable to capture the difference in transition functions $f^{(i)}$, rather than underlying $z^{(i)}_0$, we use the same initial condition $x^{(i)}_0 = 0, \forall i$. We generate data for $t\in[0, 1.5]$, and randomly sample 10 time points for each individual data entry as observed data. 

\begin{figure}
    \centering
    \includegraphics[width=0.7\columnwidth]{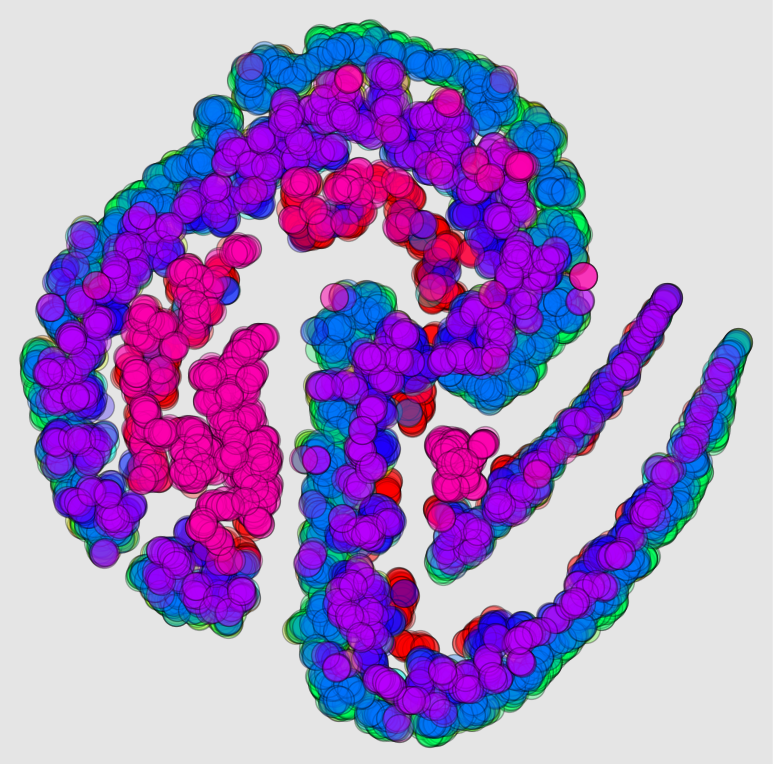}
    \caption{T-SNE representation of learned by our framework $\gamma^{(i)}$, which correspond to embedding of transition function $f^{(i)}$ of corresponding data trajectories. Each color correspond to a different value of amplitude $A^{(i)}$.}
    \label{fig:ch5_tsne_synthetic}
\end{figure}

\paragraph{Learning a meaningful representation of transition function.}
Studying the form of equation \eqref{eq:ch5_synthe_exp}, it is clear that samples of $x^{(i)}_t$ are separated corresponding to different values of $A^{(i)}$. To illustrate the ability of our model to learn meaningful representation of transition functions $f^{(i)}$, we plot T-SNE representation of learned by our framework $\gamma^{(i)}$, Figure~\ref{fig:ch5_tsne_synthetic}. While there is some overlap between classes of derived $gamma$s corresponding to different values of amplitude $A$, there is a clear separation as one would expect by observing equation ~\eqref{eq:ch5_synthe_exp}. A good separation between encoded values $\gamma$ of temporal trajectories allows us to perform statistical inference, including sampling, which we are going to talk about next.

\paragraph{Variational sampling of temporal trajectories.}
Recall, we described how we can correct posterior inaccuracies of learned approximate posterior distribution of $\gamma^{(i)}$, which allows us to perform variational sampling of high quality. Namely, for each sample $x^j$ from the training data, we encode parameters $(\mu_\gamma^j, \Sigma_\gamma^j)$ and draw $n_\gamma$ independent samples of $\gamma^j \sim N(\mu_\gamma^j, \Sigma_\gamma^j)$. Thus, given $N$ training samples, we obtain $N \times n_\gamma$ samples of $\gamma$, using which a GMM is learned. 

\begin{figure}
    \centering
    \includegraphics[width=1\columnwidth, trim={0.9cm, 0cm, 0cm, 1cm}, clip]{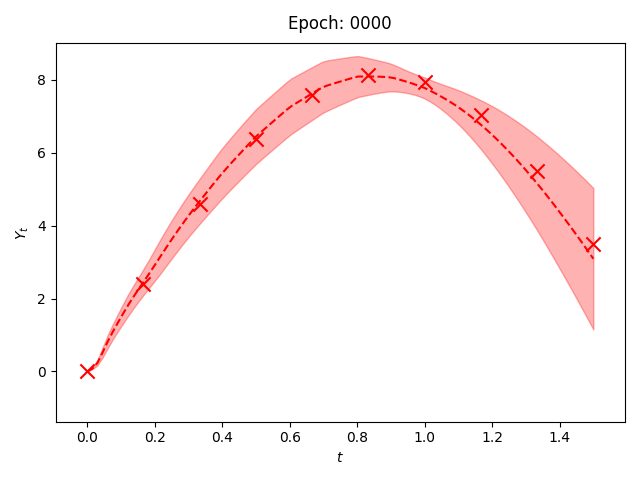}
    \caption{We demonstrate the variational sampling of temporal trajectories, given the 10 samples of observed data (x) used to train the model to derive representation of transition function. Red region is 95\% credible interval and dashed line corresponds to the posterior mean.}
    \label{fig:ch5_samples_synthetic}
\end{figure}

Given a learned GMM, we are able to sample temporal trajectories corresponding to data $x^{(i)}$, which is presented in Figure~\ref{fig:ch5_samples_synthetic}. In addition to observed values used for training the model (crosses on the image), we represent the 95\% credible interval (red region) and a posterior mean (red dashed line), which are derived through sampling  $\gamma^{(i)}$ from learned GMM and passing it through decoder, to obtain new values $\hat{x}^{(i)}$. As we can see that all observed values lie inside the credible interval, which indicate a simple check of our model's capability to generate variational samples of temporal trajectories. Next, we are going to discuss more of a statistical inference, and demonstrate the usability of our framework to detect trajectories outside the distribution.

\paragraph{Out-of-Distribution (OOD) detection.}
To evaluate the ability of our model to detect OOD trajectories, we introduce another set of trajectories, defined as: 
\begin{equation}
\left[
\begin{array}{l}
x^{B,(i)}_t = \sin(\psi^{(i)}_t) +\varepsilon^{(i)}\\
\frac{\psi^{(i)}_t}{dt} = 2\pi \cdot B^{(i)}\\
\varepsilon^{(i)}\sim N(0,10^{-3})
\end{array}
\right.
\label{eq:ch5_synthe_exp2}
\end{equation}

Similar to the previous setup, $x^{B,(i)}_t$ conceptually might represent the $i$ patient's observation at time $t$, but belongs to the group with an underlying biological risk factor, which effects the development of the temporal process $x^{B,(i)}_t$. in this case the frequency coefficient $B^{(i)}$ is defined per group $i$, e.g. patient, and can vary among groups.

We generate $10,000$ examples with $10$ unique values of $B^{(i)}\sim Unif(0, 10)$, with the same initial condition $x^{B,(i)}_0 = 0, \forall i$. We generate data for $t\in[0, 1.5]$, and randomly sample 10 time points for each individual data entry as observed data.

\begin{figure}
    \centering 
    \includegraphics[width=0.35\columnwidth, trim={0cm, 0cm, 0cm, 0cm}, clip]{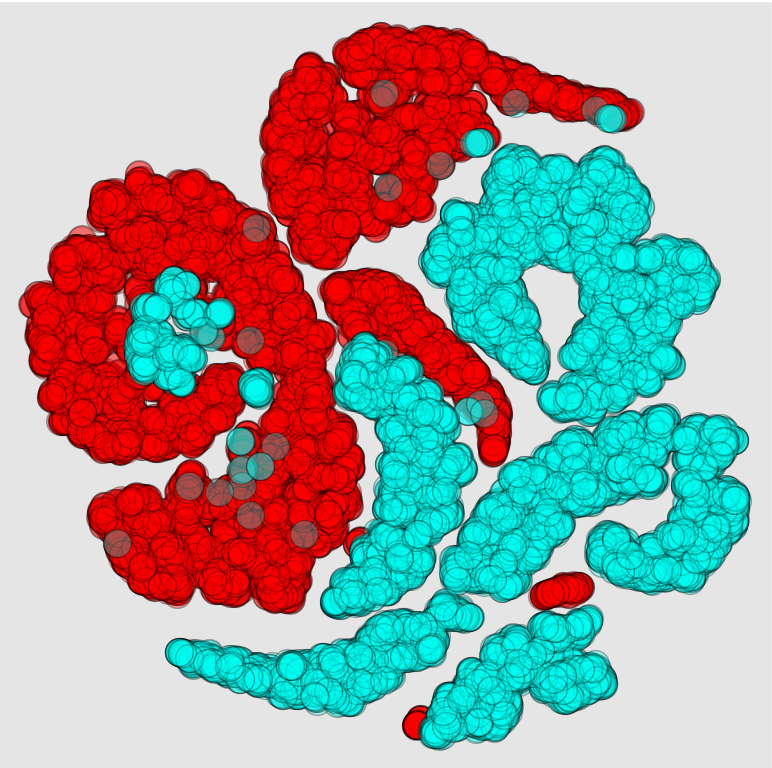}
    \includegraphics[width=0.5\columnwidth]{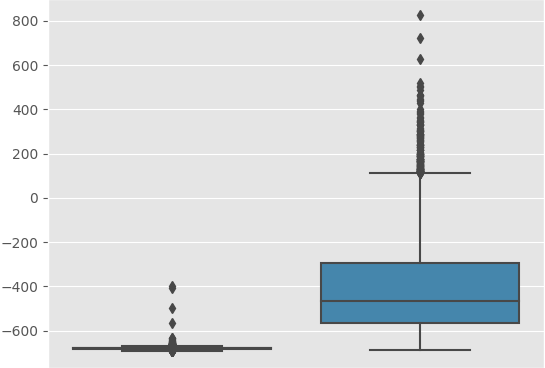}
    \caption{Outlier test: red is set participating in the training, blue is newly observed set $B$. Noticed the meaningfull separation in T-SNE representation of embedding between two distributions. in addition, we provide boxplots of negative log-likelihood for corresponding samples. As we can see, the negative log-likelihood of data for the set B is much higher, then for a training set, which clearly indicates the outlier nature of set B.}
\label{fig:ch5_ood_synth}
\end{figure}

Recall, we describe the ability of our model to conduct OOD. Given a set of observed data $x^{(i)}$, we train our model to learn the approximate posterior distribution (and corresponding GMM) of $\gamma^{(i)}$ -- the latent representation of transition functions corresponding to $x^{(i)}$. Then, given a trained model, we embed newly observed trajectories $x^{B, (i)}$ to the $\gamma^{B(i)}$ and conduct the likelihood outlier test to understand whether  $\gamma^{B(i)}$ belongs to the learned GMM -- distribution of $\gamma^{(i)}$. In Figure~\ref{fig:ch5_ood_synth} we present two findings: on the left side we have a T-SNE representation of embedding of trajectories and on the right, the boxplot of negative log-likelihood for corresponding samples. As we can see, the negative log-likelihood of data for the set B is much higher than for a training set, which clearly indicates the outlier nature of set $B$.

In the following subsection we provide an in-depth exploration of variational sampling of trajectories and statistical inference techniques, which is performed on more complex computer vision data sets, demonstrating the versatility of our framework.

\subsection{Pittsburgh Compound B or PIB data}
To show the ability of our model to do a variational sampling of temporal trajectories with limited number of  samples, we consider Pittsburgh Compound B data \cite{betthauser2022multi}. Data contains 314  records of patients, with each record up to 6 time steps. PIB data provides information about 118 regions of brain scans, patients age at visit and has a property of irregular time series. 

\begin{figure}
    \centering 
    \includegraphics[width=1\columnwidth]{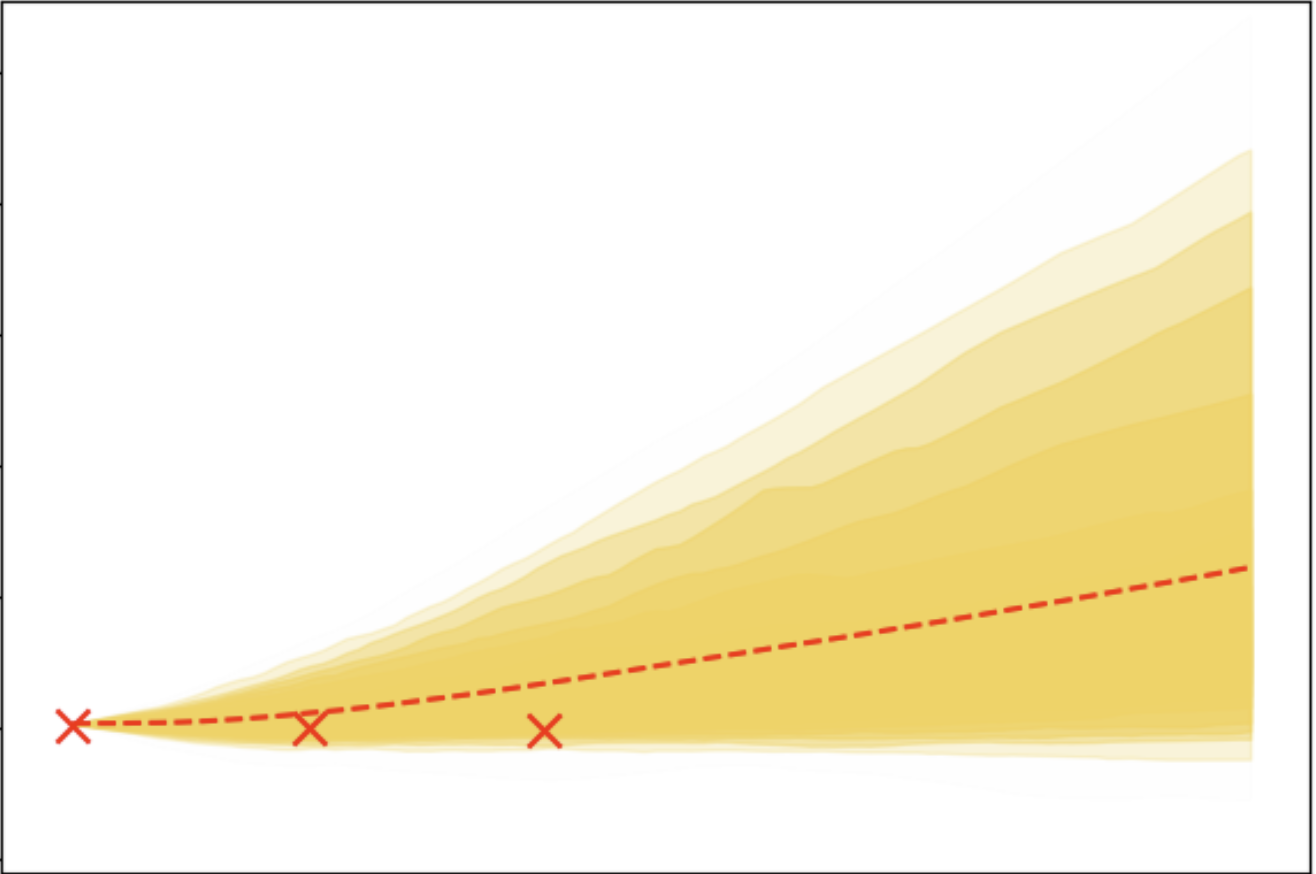}
    \caption{Variational sampling of temporal trajectories of one of the 118 brain regions for 6 time steps, given only 3 time steps observed per patient. Yellow lines are generated trajectories, red dashed line is a posterior average, and red crosses are observed time steps.}
\label{fig:pib}
\end{figure}

In Figure~\ref{fig:pib} we provide an example of variational sampling of temporal trajectories of one of the 118 brain regions for 6 time steps, given only 3 time steps observed per patient.

\subsection{Rotating MNIST}
\begin{figure}
    \centering 
  \includegraphics[width=1\columnwidth]{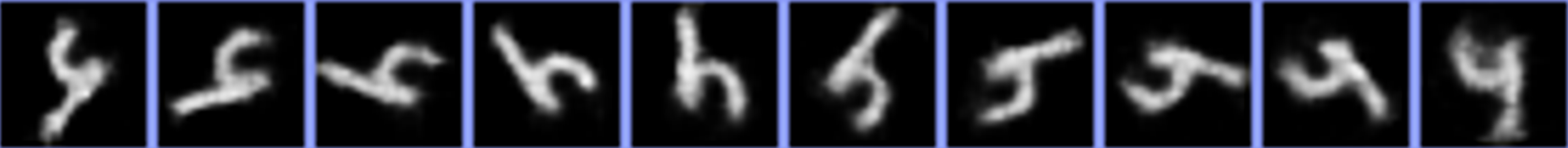}
  \includegraphics[width=1\columnwidth]{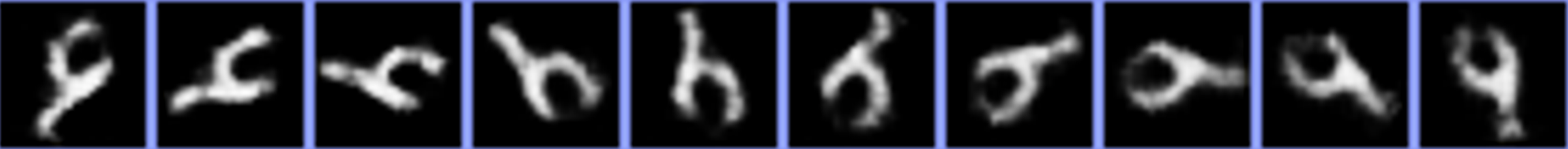}
  \includegraphics[width=1\columnwidth]{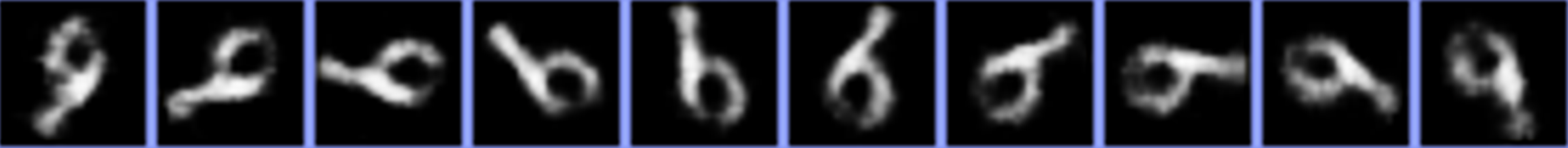}
  \includegraphics[width=1\columnwidth]{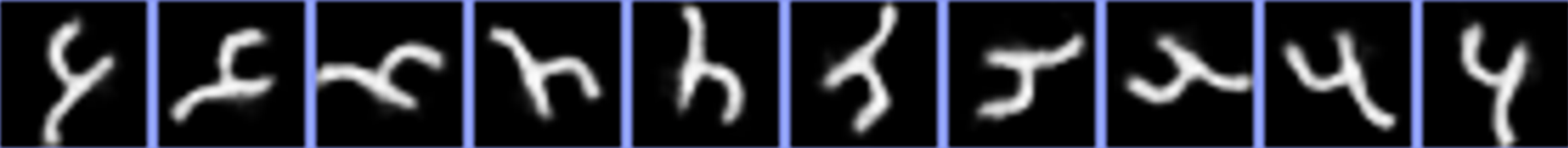}
  
  \smallskip
  
  \includegraphics[width=1\columnwidth]{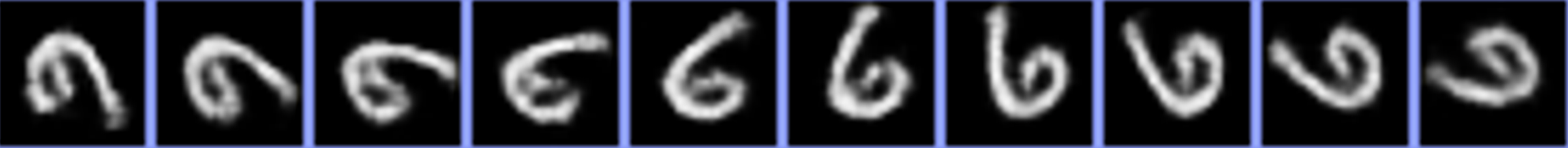}
  \includegraphics[width=1\columnwidth]{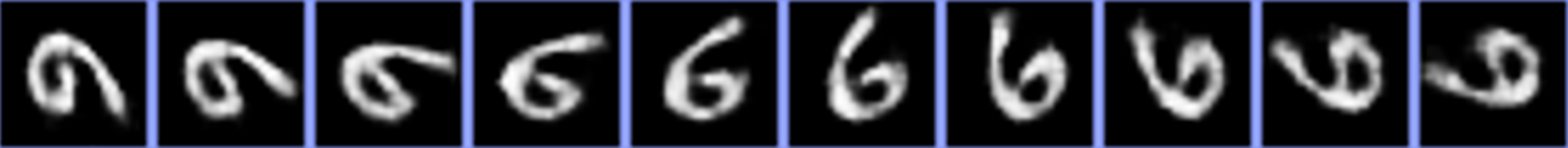}
  \includegraphics[width=1\columnwidth]{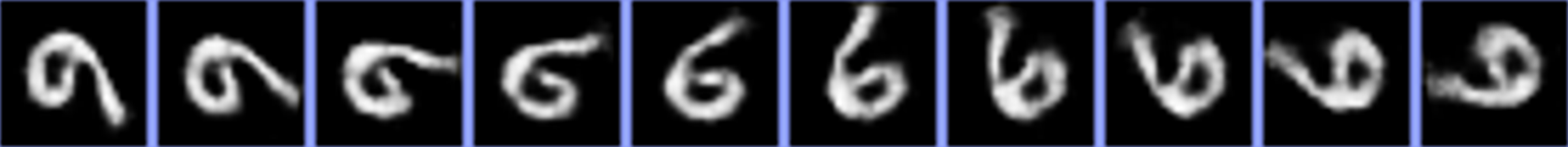}
  \includegraphics[width=1\columnwidth]{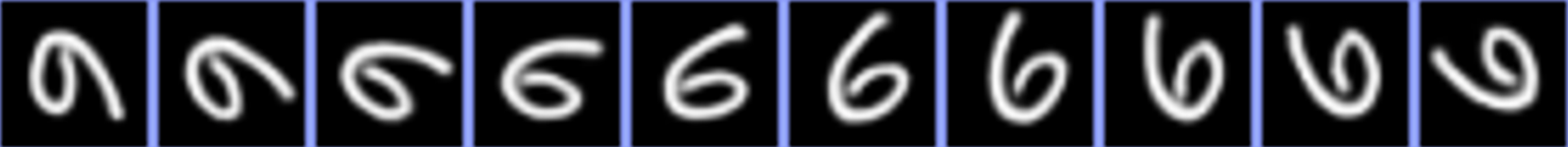}

  \caption{Reconstruction of the same samples, from top to bottom by block: VAE-RNN, NODE, BNN-NODE, FODE (our). Clearly, our model provides the most clear reconstruction.
  \label{fig:rotmnist_samples}}
\end{figure}

In this section we provide additional samples of trajectories. 
To show the quality of our reconstruction, regardless of MSE values from the table in main paper, in Figure~\ref{fig:rotmnist_samples} we present the ablation studies of several samples for different baselines.

In addition to comparison to quality of reconstruction, we show the variation of samples derived by sampling different representations of the transition function $\gamma$, but preserving the initial value of the latent representation $z_0$. 

\begin{figure}
    \centering 
  \includegraphics[width=1\columnwidth]{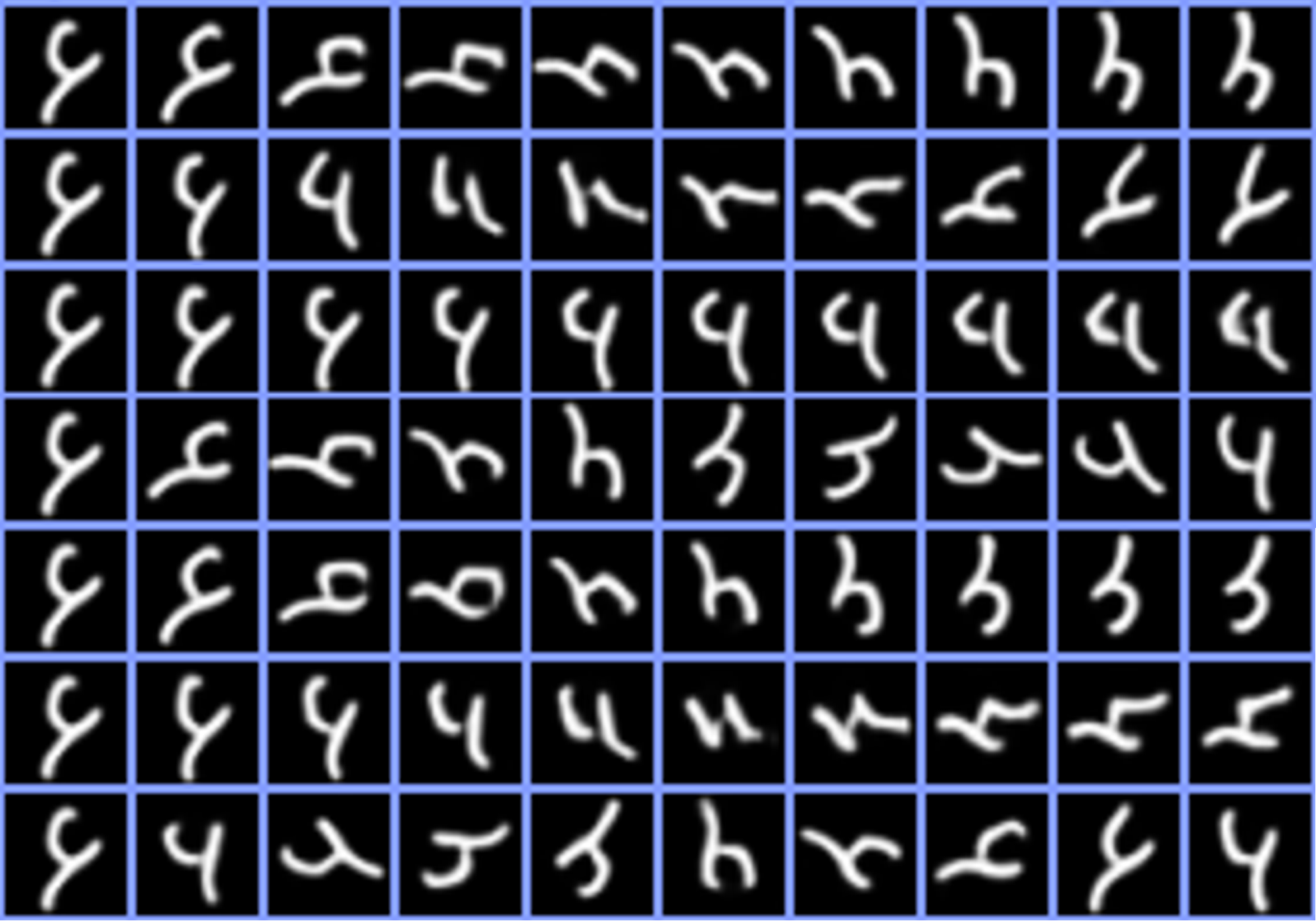}
  
  \caption{Variation of samples for our model. Notice the difference in the angle of rotation given the same initial value, but varying the representation of transition function $\gamma$.
  \label{fig:rotmnist_variation}}
\end{figure}

\subsection{MuJoCo}
Below we provide several samples from our model for different sets of MuJoCo, namely Hopper, Walker and 3-poles cart. Notice how trajectories are different, despite the same initial values.
\begin{figure}
    \centering 
  \includegraphics[width=1\columnwidth]{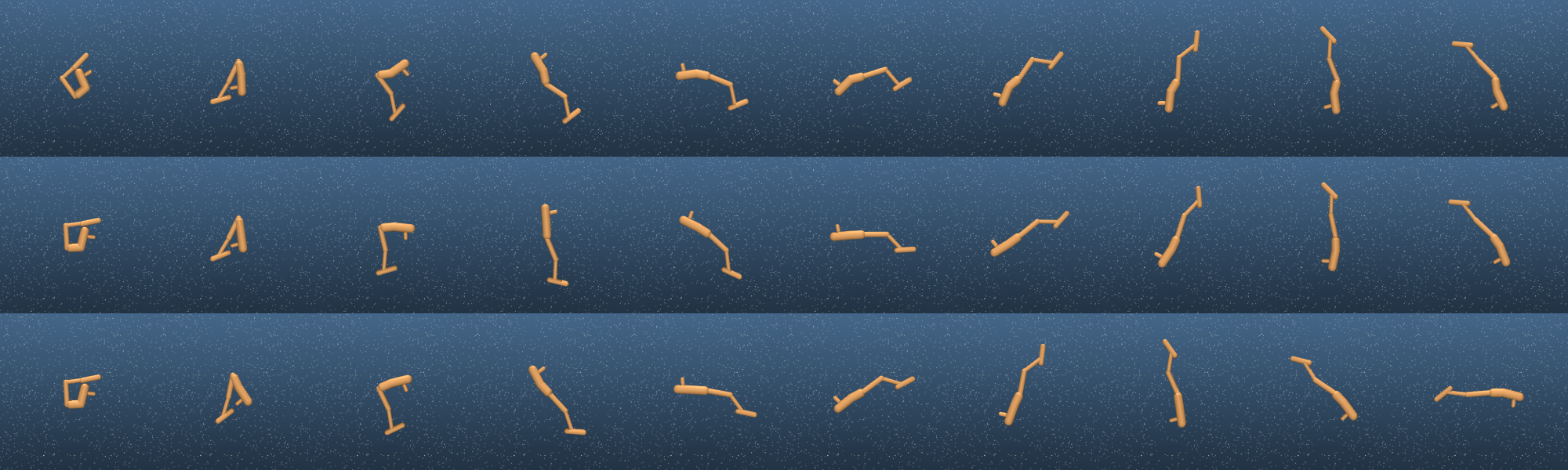}
\end{figure}
\begin{figure}
    \centering 
  \includegraphics[width=1\columnwidth]{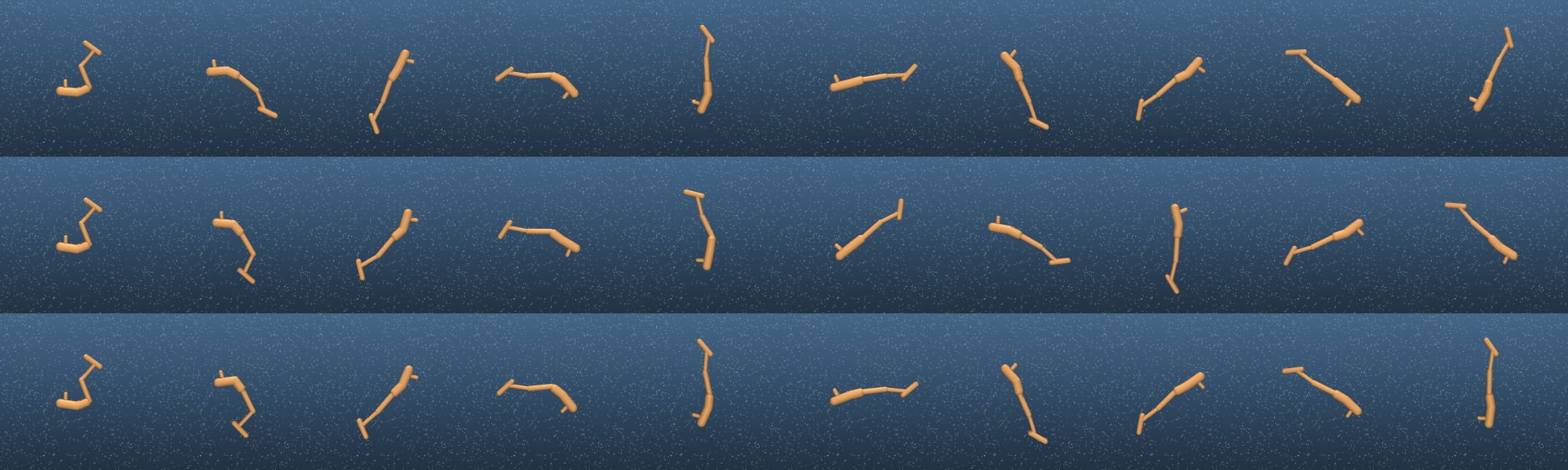}
\end{figure}
\begin{figure}
    \centering 
  \includegraphics[width=1\columnwidth]{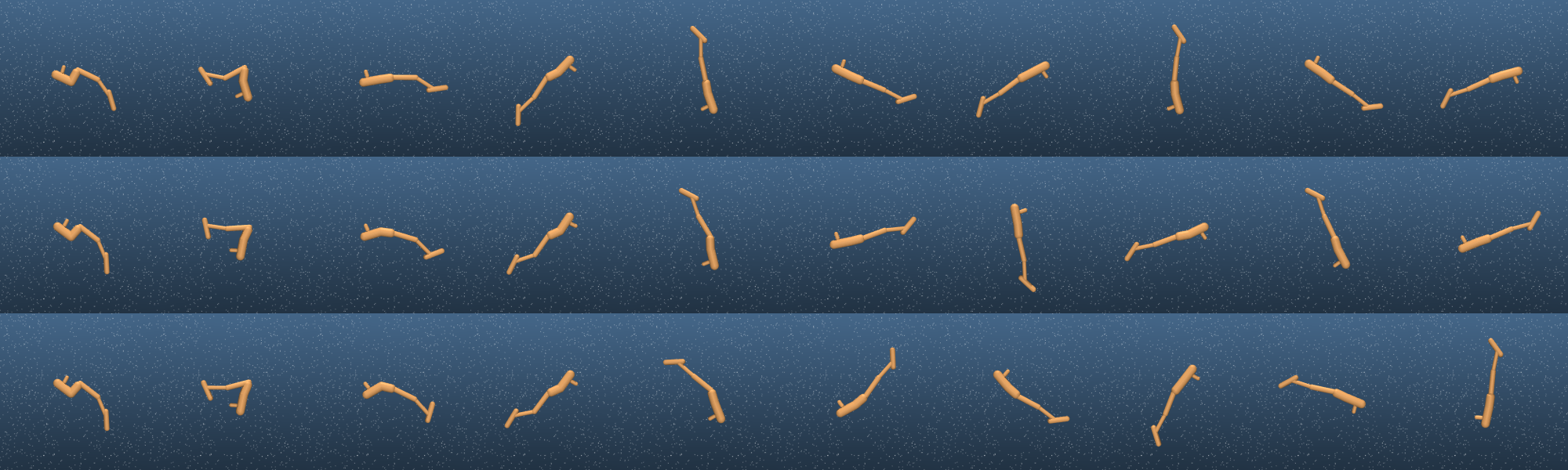}
\end{figure}

\begin{figure}
    \centering 
  \includegraphics[width=1\columnwidth]{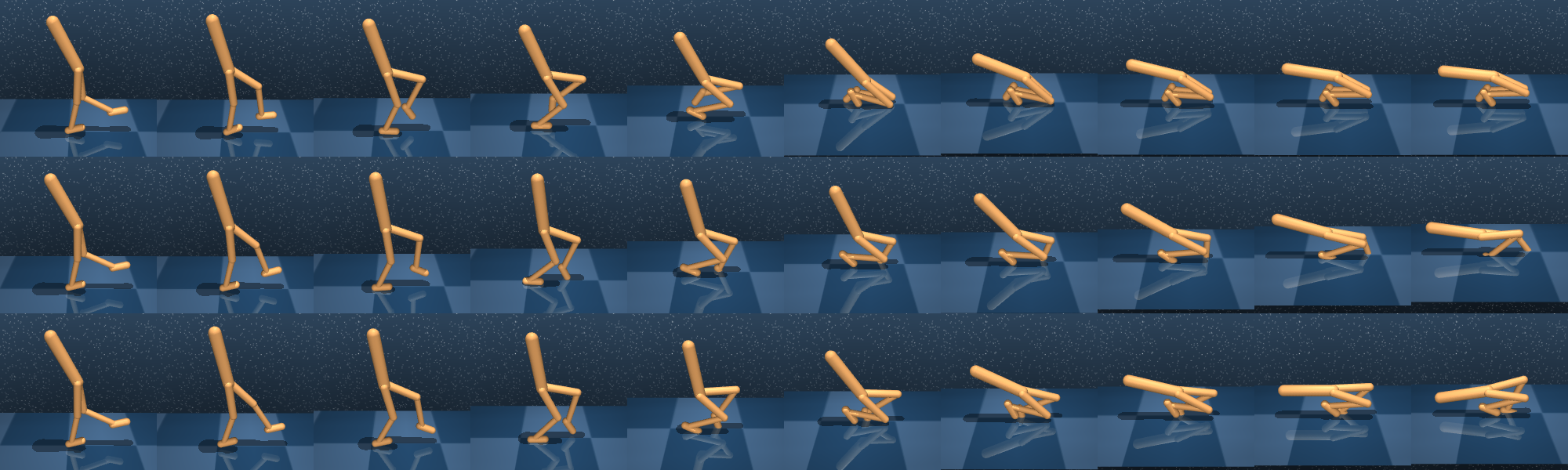}
\end{figure}
\begin{figure}
    \centering 
  \includegraphics[width=1\columnwidth]{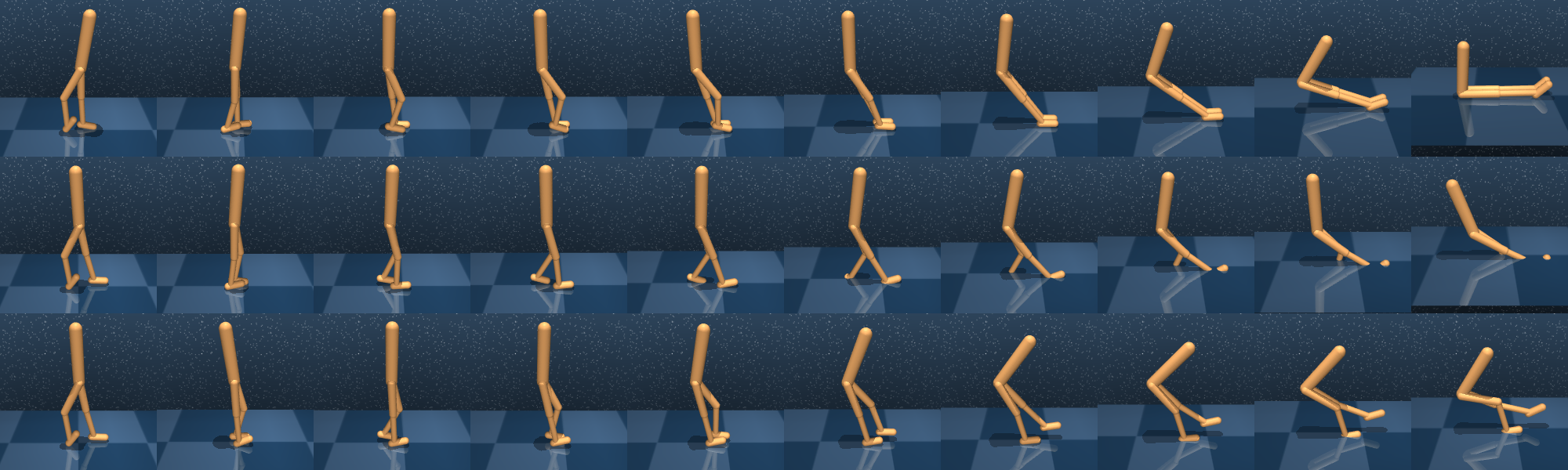}
\end{figure}
\begin{figure}
    \centering 
  \includegraphics[width=1\columnwidth]{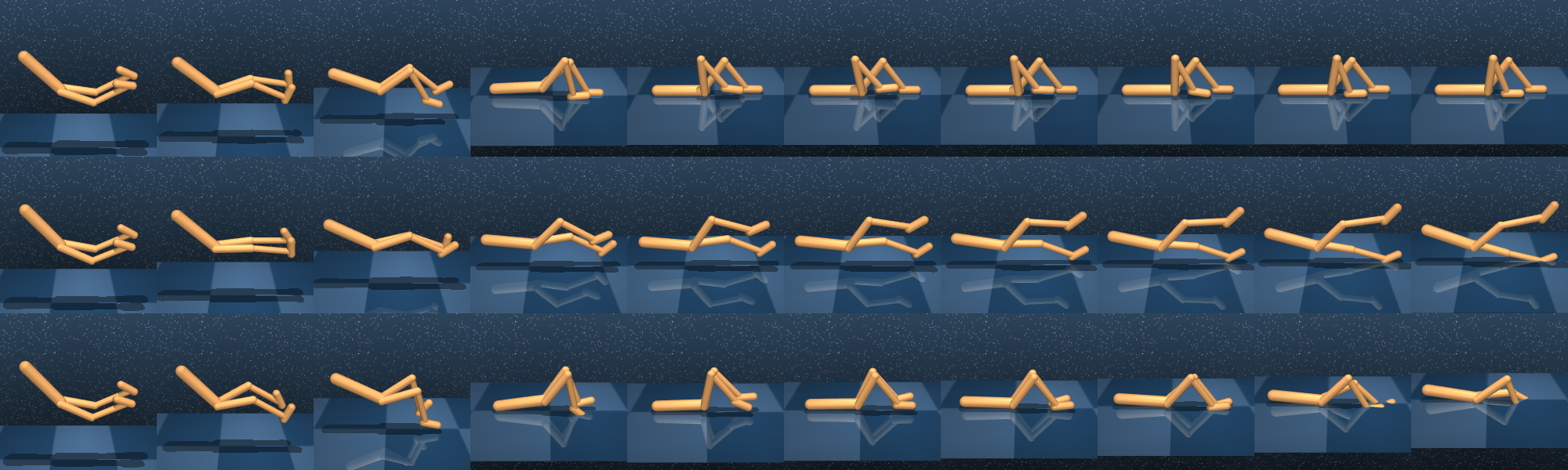}
\end{figure}

\begin{figure}
    \centering 
  \includegraphics[width=1\columnwidth]{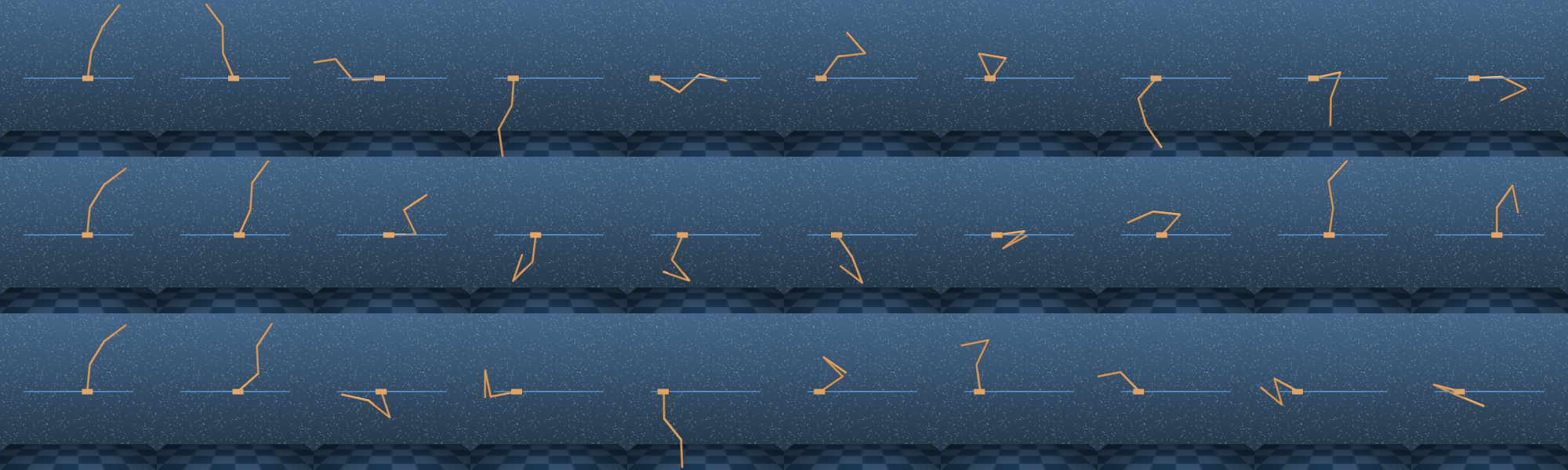}
\end{figure}
\begin{figure}
    \centering 
  \includegraphics[width=1\columnwidth]{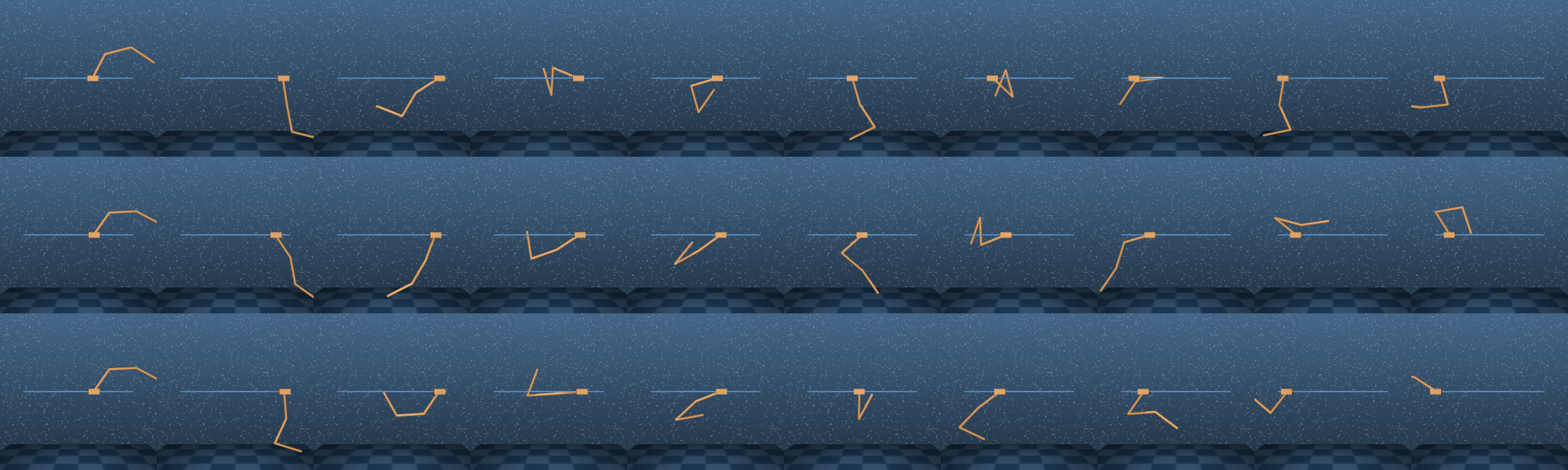}
\end{figure}
\begin{figure}
    \centering 
  \includegraphics[width=1\columnwidth]{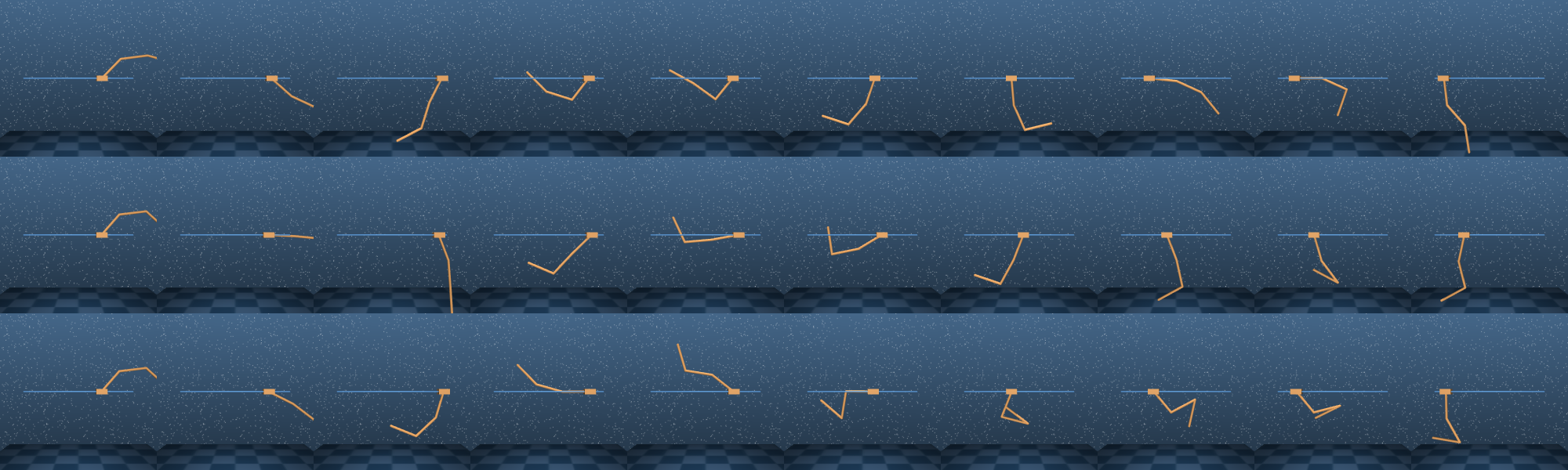}
\end{figure}

{\small
\setcitestyle{numbers}
\bibliography{refs}
}

\end{document}